\documentclass[runningheads]{llncs}

\usepackage[final,year=2026,ID=\textbf{979} ]{eccv}

\usepackage{eccvabbrv}

\usepackage{graphicx}
\usepackage{booktabs}
\usepackage{array}
\newcolumntype{C}{>{\centering\arraybackslash}p{1.2cm}}
\usepackage{csquotes}
\usepackage{comment}
\usepackage[accsupp]{axessibility}  

\usepackage{hyperref}

\usepackage{orcidlink}

\begin{document}

\title{Segmenting Visuals With Querying Words: Language Anchors For Semi-Supervised Image Segmentation} 

\titlerunning{HVLFormer}

\author{Numair Nadeem\inst{1}  \and
Saeed Anwar\inst{2}  \and
Muhammad Hamza Asad\inst{3} 
\and
Abdul Bais\inst{1} }

\authorrunning{N.Nadeem et al.}

\institute{
University of Regina, Canada\\
\email{nng794@uregina.ca, abdul.bais@uregina.ca}
\and
The University of Western Australia, Australia\\
\email{saeed.anwar@uwa.edu.au}
\and
University Canada West, Canada\\
\email{muhammadhamza.asad@ucanwest.ca}
}

\maketitle

\begin{abstract}

Vision–Language Models (VLMs) provide rich semantic priors but are underexplored in Semi-supervised Semantic Segmentation. Recent attempts to integrate VLMs to inject high-level semantics overlook the semantic misalignment between visual and textual representations that arises from using domain-invariant text embeddings without adapting them to dataset- and image-specific contexts. This lack of domain awareness, coupled with limited annotations, weakens the model’s semantic understanding by preventing effective vision-language alignment. As a result, the model struggles with contextual reasoning, shows weak intra-class discrimination, and confuses similar classes. To address these challenges, we propose the Hierarchical Vision–Language Transformer (HVLFormer), which achieves domain-aware and domain-robust alignment between visual and textual representations within a query-driven mask-transformer architecture. Firstly, we transform text embeddings from a pre-trained VLM into multi-scale, dataset-aware textual object queries that capture class semantics from coarse to fine granularity and enhance the model's semantic understanding. Next, these queries are refined using image-specific visual context, aligning global textual semantics with local scene structures and improving class discrimination. Finally, to achieve domain-robustness, we introduce cross-view and modal consistency regularization, which enforces prediction consistency within the mask-transformer architecture across augmented views. It ensures that language queries remain robust to intra-class variations and diverse visual scenes without overfitting to the small labeled set, while maintaining stable vision–language alignment under perturbations during decoding. With less than 1\% training data, HVLFormer outperforms state-of-the-art methods on Pascal VOC, COCO, ADE20K, and Cityscapes. Our code and results will be available on GitHub.

\end{abstract}

\section{Introduction}
\label{sec:intro}
Developing semantic segmentation models that require minimal labeled data remains a significant challenge in computer vision, particularly in autonomous driving, agriculture, and medical imaging, where acquiring dense pixel-level annotations is expensive and time-consuming. Semi-supervised Semantic Segmentation (SSS) addresses this challenge by leveraging a small subset of labeled images alongside a large pool of unlabeled data, aiming to achieve strong performance without full supervision~\cite{Nadeem_2025_CVPR,10.1007/978-3-031-91835-3_19,Zhao2024Alternate,Chi2024Adaptive,Yuan2024Dynamically}. Typical strategies include adversarial training~\cite{mittal2019semi}, and self-training~\cite{souly2017semi,yang2022st++,yang2023revisiting}, which aim to extract supervisory signals from unlabeled data. 
Despite recent advances in achieving precise segmentation~\cite{10839097,hoyer2024semivl}, SSS models still frequently misclassify, particularly in scenarios with significant intra-class variation, visually similar classes, and rare classes. For instance, confusing sofas with chairs often arises from the limited representation of rare categories within the small labeled subset~\cite{zhong2023cvpr}. 
Unlike unsupervised domain adaptation and domain generalization, which benefit from fully labeled source data to establish a strong semantic understanding, SSS lacks sufficient labeled examples, resulting in weak feature learning and limited semantic understanding. 
Recent advances in VLMs~\cite{fang2023eva, liu2023llava,fang2023eva} offer a promising direction, as they encode domain-invariant semantic representations that align visual and textual modalities within a shared embedding space~\cite{abdelfattah2023cdul}. Motivated by this, recent works integrated VLM-derived text embeddings to guide pixel-level predictions. However, their role in SSS remains underexplored, as most VLM-based segmentation methods either operate without dense annotations~\cite{Cha2023,Xu2022,Zhou2022}, resulting in limited segmentation precision, or rely on large-scale labeled datasets~\cite{Ding2022,Xu2023,Zhou2023}, which are expensive. SemiVL~\cite{hoyer2024semivl} partially bridges this gap by adapting CLIP~\cite{Radford_2021_ICML} text embeddings through spatial fine-tuning and a language-aware decoder. In parallel, Pseudo-SD~\cite{zhao2025pseudo} leverages pseudo-labels and text embeddings to fine-tune stable diffusion models for generating labeled synthetic images.

The approaches mentioned above pose two key issues. First, while they emphasize domain invariance in text embeddings to provide global class semantics, they overlook domain awareness—failing to effectively adapt text embeddings to dataset- and image-specific contexts that are crucial for compensating for weak semantic understanding in low-data regimes. Trained on generic web-scale corpora, pre-trained VLM embeddings encode broad semantics that miss the fine contextual cues necessary for precise feature learning. For example, \textit{\enquote{chair}} and \textit{\enquote{sofa}} share overlapping meanings in generic corpora. However, within the ADE20K~\cite{ade20k} dataset, they differ in shape and spatial context: chairs often appear around tables, whereas sofas are typically located in living areas. Without adapting text embeddings to such domain-specific semantics, the model learns blurred decision boundaries, struggles to separate correlated classes, and fails to establish stable representations for under-represented categories. Second, as these methods are predominantly optimized through pixel-level supervision, the language component remains a weak auxiliary cue rather than an actively aligned semantic guide, resulting in shallow vision–language alignment that hinders contextual reasoning and limits semantic knowledge extraction from pretrained VLMs. 

To address these challenges, we propose the Hierarchical Vision–Language transFormer (HVLFormer), a unified mask-transformer-based framework where language queries predict class-specific segmentation masks under limited supervision. HVLFormer utilizes domain-invariant text embeddings from pre-trained VLM~\cite{Radford_2021_ICML} as object queries and injects domain-awareness by conditioning them on dataset and image-specific contextual cues for effective pretrained VLM's knowledge extraction and multimodal alignment. To achieve domain robustness, it enforces prediction consistency across augmented views, stabilizing vision–language alignment. This design enables the model to perform explicit language-guided segmentation, enhancing contextual reasoning, class discrimination, and feature learning under limited supervision.

Our HVLFormer introduces \textit{Hierarchical Textual Query Generation (HTQG)} module that transforms VLM-derived per-class embeddings into multi-scale, dataset-aware textual queries that capture class semantics from coarse to fine granularity, thereby enhancing the model's immunity to intra-class variations and the misclassification of rare and similar classes. It aims to better capture dataset-dependent visual patterns and contextual relationships. HTQG further weighs queries corresponding to the probability of their represented class's presence in each image, reducing error propagation through image-irrelevant queries . Next, our \textit{Pixel–Text Refinement Module (PTRM)} enriches these queries by injecting image context directly from pixel features generated by a pixel decoder, aligning textual semantics with local scene structure. This refinement makes the queries more discriminative and spatially aware, allowing them to adapt to complex visual distributions and better distinguish similar classes under limited annotation. The refined textual queries then guide segmentation by interacting with pixel features in the mask-transformer decoder to form coherent semantic groups and predict segmentation masks. To enhance domain robustness and maintain stable vision–language alignment during decoding, we introduce \textit{Cross-View and Modal Consistency Regularization (CMCR)}, which enforces prediction consistency across augmented views and aligns pixel–text interactions within the transformer decoder. It is applied at every decoder layer to prevent error propagation, making queries adapt to diverse image contexts under limited supervision. This marks the first consistency-driven refinement within a mask-transformer framework.

Our main contributions are summarized as follows:
\begin{itemize}
\item We introduce a language-driven SSS framework that explicitly leverages text embeddings from pre-trained VLMs as object queries and refines them with dataset- and image-specific context. This design addresses the persistent issue of weak feature learning and semantic understanding in SSS by introducing domain-aware textual queries that adapt global semantic priors to local visual distributions under limited supervision, while a multimodal regularization loss ensures stable vision–language alignment during training.
\item We develop a unified hierarchical vision–language architecture that enables rich interactions between domain-aware textual queries and multi-scale pixel features. Through progressive vision–language alignment, framework bridges global linguistic semantics with dataset-specific visual representations. Following this alignment, the domain-aware textual queries inject rich semantic knowledge into the model, compensating for its weak semantic understanding. This process, reinforced by cross-view and modal consistency regularization, leads to stronger class discrimination, improved contextual stability, and superior label efficiency in low-annotation regimes.

\end{itemize}

\begin{figure*}[!tbp]
\centering
\includegraphics[width=0.98\textwidth]{images/HVLFormer-ECCV.png}         
\caption{Overview of HVLFormer:
The pre-trained VLM encodes visual and textual features, generating dataset-aware text embeddings from class names and dataset attributes. The \textbf{HQG} module transforms them into multi-scale queries, with \textbf{SRE} weighing queries based on the presence of their corresponding classes in an image, refined by \textbf{PTRM} through pixel–text alignment. These queries guide the transformer decoder to predict class-specific masks, with each query responsible for generating a segmentation mask of its represented class, while \textbf{CMCR} enforces cross-modal consistency for stable vision–language alignment under limited supervision. }
\label{fig:multi_resolution_training}
\vspace{-4mm}
\end{figure*}
\section{Related Works}

\subsection{\textbf{Semi-supervised Semantic Segmentation (SSS)}}
Early methods adopted adversarial frameworks~\cite{Chen2021Semi,Wu2023Querying} to extract supervisory signals from unlabeled data. However, these approaches often suffered from instability~\cite{10839097}. Subsequent work simplified the paradigm through entropy minimization~\cite{ijcai2021pseudoseg} and consistency regularization~\cite{Guan2022Unbiased,Yang2023Shrinking}, which have become the foundation of modern approaches. Utilizing them, self-training strategies~\cite{hoyer2024semivl} generate pseudo-labels for unlabeled data and refine them iteratively while enforcing prediction consistency under augmentations such as CutOut and CutMix~\cite{Yun2019CutMix}. Similarly, cross pseudo supervision~\cite{Chen2021Semi} further improved stability through co-training, and adaptive equalization Learning~\cite{Hu2021Semi} biased CutMix towards under-represented classes. To address poor early-stage pseudo-label quality, ST++~\cite{yang2022st++} introduced staged pseudo-label augmentation to enable progressive refinement. Recently, methods like UniMatch \cite{Yang_2023_CVPR} have adopted confidence-based filtering~\cite{10839097,Zhao2023Instance}, inspired by FixMatch~\cite{Sohn2020FixMatch}, in which weak–strong augmentation pairs enforce consistency. These designs influenced vision–language extensions such as SemiVL~\cite{hoyer2024semivl}, which incorporates CLIP~\cite{radford2021learning} text embeddings as semantic regularizers. However, SemiVL~\cite{hoyer2024semivl} treats textual cues as single-level, domain-invariant embeddings, limiting their capacity to capture intra-class diversity and fine-grained semantics. In contrast, HVLFormer enhances the textual embeddings by converting them into domain-aware, multi-scale per-class queries that effectively adapt to local visual context. These refined queries serve as dynamic semantic anchors, improving same-class pixel grouping and multimodal alignment, with the proposed CMCR further enforcing stable multimodal alignment under sparse supervision.

\subsection{Language-Driven Segmentation Frameworks }

VLMs use large-scale image–text pretraining to construct a shared embedding space between visual and textual modalities, enabling generalization across diverse tasks such as image classification, retrieval, and open-vocabulary recognition. Extending the mentioned semantic segmentation paradigm, preliminary approaches such as MaskCLIP~\cite{dong2023maskclip} and GroupViT~\cite{Xu2022} attempted to learn segmentation solely from image–caption pairs, without pixel-level annotations, and demonstrated that segmentation cues can emerge from contrastively trained encoders; however, their reliance on image–caption supervision produced coarse as well as noisy masks. 
To improve mask quality, ZegCLIP~\cite{Zhou2023} aligns dense visual features with textual embeddings via learned decoders and attention mechanisms. Meanwhile, parameter-efficient tuning approaches, such as prompt tuning~\cite{zhou2022coop, jia2022visualprompt}, improve generalization with minimal supervision. However, both contrastive pretraining and prompt-tuning strategies remain limited by their reliance on coarse global semantics, lacking the fine-grained, image-specific alignment needed for precise dense prediction.

Lately, query-based transformer architectures have achieved strong performance in dense prediction. Mask2Former~\cite{cheng2022masked} aggregates multi-scale visual features from the vision encoder into spatially rich pixel features through the pixel decoder. Randomly initialized object queries interact with these features through masked cross-attention in a transformer decoder to group pixels of the same class.
Building on this, TQDM~\cite{pak2024textual} introduces a textual-query-driven framework for domain-generalized segmentation, utilizing CLIP-based text embeddings to initialize object queries and enhance the semantic clarity of pixel features by computing text-to-pixel attention before the transformer decoder. However, the one-way text-to-pixel attention restricts feedback from the visual to the textual space, preventing queries from adapting to image-specific variations before being fed into the transformer decoder. The TQDM’s focus on domain invariance is effective for domain generalization but ignores domain awareness—an essential factor in SSS. As a result, TQDM~\cite{pak2024textual} adapts poorly to queries under SSS, widening the modality gap because generic textual semantics fail to anchor weak visual features, leading to unstable feature correspondence.

Moreover, the quality of these initial queries to the decoder is critical, as they determine how effectively the decoder groups pixels and propagates semantics across layers. Generic, domain-invariant queries such as \textit{“bus”} and \textit{“train”}—which share a broad “transport” concept but differ contextually in cityscapes~\cite{cityscapes}—lack the fine-grained cues required for accurate segmentation under limited training examples. In contrast, domain-aware queries provide stronger semantic grounding, enabling clearer and stable pixel grouping during decoding. 
HVLFormer addresses this by introducing domain-awareness during text embedding generation through dataset-specific prompting and by injecting visual context from pixel features into these queries. However, without explicit regularization, the scarcity of labeled data limits supervision signals, causing vision–language misalignments to amplify across decoder layers and degrade semantic consistency. To counter this, we enforce layer-wise consistency regularization, which, through augmented visual–language interactions, forces the model to effectively enrich textual queries with image-specific context, ensuring coherent mask and class predictions while maintaining stable vision–language alignment throughout decoding.

\section{HVLFormer }


HVLFormer, a vision-language segmentation framework, is built on a mask-classification backbone~\cite{cheng2022masked}, with the pre-trained VLM's image encoder $\mathbf{E}_I$ that is fine-tuned on labeled images using standard cross-entropy loss and a text encoder $\mathbf{E}_T$ that remains frozen to preserve the VLM's semantic space. For an input image $\mathbf{I} \in \mathbb{R}^{H \times W \times 3}$, $\mathbf{E}_I$ produces multi-scale feature maps that are fed to the HVLFormer pixel decoder, providing multi-scale pixel-features $z:\{z^e\}_{e=1}^E$ and dense per-pixel embeddings $\mathbf{Z}_{pixel} \in\mathbb{R}^{H \times W \times D}$. The textual queries from text embeddings, along with pixel features, are input to a transformer decoder to generate segmentation masks.

To effectively align the pixel features with language semantics (textual queries), HVLFormer introduces a series of vision–language interaction modules. The HTQG module generates multi-scale, dataset-aware per-class textual queries, which are enriched by image-context in PTRM through cross-modal spatial interactions, and then produces segmentation masks after being processed by the transformer decoder~\cite{cheng2022masked}. Finally, CMCR enables domain-robust vision-language alignment through consistency regularization, thus forcing the model to effectively enrich textual queries with diverse image contexts. The overall pipeline is illustrated in Fig.~\ref{fig:multi_resolution_training}.

\subsection{Hierarchical Textual Query Generation (HTQG)}
 
Learnable prompting architectures have demonstrated a strong ability to generate semantically coherent prompts~\cite{hoyer2024semivl}. Building on this capability, we employ a learnable dataset-specific prompt $p_k$ to produce attribute-rich textual descriptions for each semantic class. The prompt $p_k$ takes the \textit{dataset attributes} with \textit{class name} as inputs, ensuring that the generated text captures scene-specific and compositional semantics relevant to the dataset. This design is important because the visual appearance of a class can vary significantly across datasets depending on their application context. For example, in urban driving scenes (Cityscapes~\cite{cityscapes}), classes such as \enquote{car} and \enquote{road} are typically observed from a street-level perspective with strong geometric regularities, whereas in object-centric images (Pascal VOC~\cite{pascal_voc}), the same classes appear under diverse backgrounds, scales, and viewpoints. By incorporating dataset attributes, the text encoder $E_T$ generates dataset-aware textual embeddings that better align with the visual distributions of each dataset. Details of the attributes used for each dataset are provided in the supplementary material.

For each class $k$, dataset aware textual prompt is encoded using the frozen text encoder $E_T$, i.e. $\mathbf{t}_k =   E_T\left(p_k\right) \in \mathbb{R}^C.$

However, in dense segmentation, single-level text embeddings fail to capture the full range of intra-class variations, an issue further amplified by limited labeled data. They struggle to represent both coarse structure and fine local details, leading to poor generalization across diverse visual contexts. Moreover, applying all class-level queries to every image introduces noise from absent classes. For example, a query corresponding to a class such as \textit{bus} may attempt to attend to bus-like patterns even when no such object exists in the image, introducing spurious responses that weaken model attention and degrade the semantic clarity of pixel features. 

To overcome these limitations, HTQG jointly performs Hierarchical Query Generation (HQG) and Semantic Relevance Estimation (SRE). HQG projects each class embedding $\mathbf{t}_k$ into multiple abstraction levels—from coarse to fine—to model semantics across scales and improve discrimination between similar or rare classes. Specifically, $\mathbf{t}_k$ is transformed by $E$ distinct MLP heads, each aligned with $e^{th}$ pixel feature $\{z^e\}_{e=1}^E$: $\hat{\mathbf{Q}} =\{\mathbf{q}_k^{e} = \text{MLP}_e(\mathbf{t}_k) \}_{k=1,e=1}^{K,E}\in \mathbb{R}^{E\times K \times  D}$ . Following the pixel decoder’s coarse-to-fine hierarchy, the heads generate queries at increasing abstraction levels of the same class rather than separate subclasses. Coarse queries encode global object structure, while finer queries capture texture and boundary details. During training, each query is refined only through its aligned feature map in PTRM, ensuring a coarse-to-fine hierarchy in queries. Furthermore, to prevent redundancy, a diversity regularization term: $\ell_{\text{div}} = \sum_{e \neq e'} \left\| \frac{(\mathbf{q}_k^{e})^\top \mathbf{q}_k^{'e}}{\|\mathbf{q}_k^{e}\| \cdot \|\mathbf{q}_k^{'e}\|} \right\|_2^2,$ encourages distinct, complementary semantics across scales.

Simultaneously, SRE estimates the likelihood of each class being present in the image, thereby suppressing irrelevant queries. For each class, a relevance score $s_k$ is computed by projecting multi-scale pixel features from the pixel decoder into a shared space with text embeddings via a lightweight adapter as: $s_k = \sigma\left(\text{MLP}\left([\phi(f_1, f_2, f_3), \mathbf{t}_k]\right)\right),$ where $\phi(f_1, f_2, f_3)$ represents fused multi-scale features from $E_I$, and $\sigma(\cdot)$ is a sigmoid activation. The MLP is trained using labeled images and a supervised loss before training. Final refined hierarchical queries are obtained as $\tilde{\mathbf{Q}} = \{\tilde{\mathbf{q}}_k^{e} = s_k \cdot \mathbf{q}_k^{e}\}_{k=1,e=1}^{K,E}.$ This yields dataset-context-aware, scale-adaptive textual queries that suppress irrelevant class semantics and encode dataset-specific features, improving vision-text alignment and ensuring robust semantic understanding under limited supervision.

\subsection{Pixel-Text Refinement Module (PTRM)}
PTRM injects image-specific context into hierarchical textual queries by integrating visual cues from pixel features $z$, such as structure, texture, and illumination, thereby enabling queries to capture scene-dependent variations. Simultaneously, class-level semantics from text enhance semantic coherence in pixel features. This bidirectional adaptation aligns textual and visual representations, enabling queries to dynamically respond to image context and yield spatially coherent, semantically consistent masks. Unlike conventional cross-attention, which relies solely on token-level similarity, PTRM introduces spatially guided conditioning that allows the image to determine where and how strongly each query should influence prediction. Consequently, each query is spatially distributed over the feature map and selectively strengthened in regions whose visual patterns match its class semantics, while being suppressed in visually irrelevant regions. 

At pixel-feature level $e$, textual queries $\tilde{\mathbf{Q}}^e: \{\tilde{\mathbf{q}}_k^{e}\}_{k=1}^{K}$ and pixel feature map $z^{e}$ are projected into a shared latent space to ensure semantic–spatial comparability, resulting in text  $T^{e}$ and visual $V^{e}$ representations. 
A shallow fusion integrates semantic and spatial priors: $F^{e} = T^{e} + V^{e}.$ This coupling creates a coarse alignment between class semantics and image-spatial structure, allowing textual information to roughly map onto visual layout cues.


To localize cross-modal importance, we compute average-pooled ($\varsigma$) representations across all feature maps:
\begin{equation}
X^{e} = [\varsigma(T^{e}), \varsigma(V^{e}), \varsigma(F^{e})].
\label{eq:AP}
\end{equation}
It highlights regions that are globally informative in either modality or their fusion. Two convolutional layers with ReLU and sigmoid activations applied to $X^{e}$ then generate spatial attention weights $W^{e}$, which are decomposed into three attention maps, i.e., text-guided $W_T^{e}$, pixel-guided $W_V^{e}$, and fused attention $W_F^{e}$. These maps jointly regulate the bidirectional adaptation between modalities:
\begin{align}
T^{'e} &= T^{e} \odot W_V^{e} \odot W_F^{e}, \\
V^{'e} &= V^{e} \odot W_T^{e} \odot W_F^{e}.
\label{eq4}
\end{align}

Here, $W_V^{e}$ injects local visual cues such as texture, structure, and illumination into textual features, enriching them with image-conditioned semantics. Conversely, $W_T^{e}$ transfers class-level priors from text to pixel features, thereby injecting semantic understanding in visual representations. Finally, $W_F^{e}$ acts as a consensus gate, emphasizing regions of mutual confidence to ensure updates occur only where textual and visual information agree. 

The final aligned textual queries and enriched pixel features are obtained by applying global average pooling to the spatially distributed text feature map $T^{'e} \in \mathbb{R}^{J\times H^{e} \times W^{e}}$, aligning them into queries $\textbf{Q}^{e} \in \mathbb{R}^{K \times D}$, and using 1x1 convolution to reproject refined image features $V^{'e}$ from the shared latent space back to the decoder's original feature domain $z_{\text{out}}^{e} \in \mathbb{R}^{H^{e} \times W^{e} \times D}$, ensuring compatibility with subsequent decoding layers. 

For all $'e'$, PTRM thus outputs enhanced pixel features $z_{\text{out}}$ enriched with class semantics and refined, context-aware textual queries $\textbf{Q} = \{\textbf{Q}^{e}\}_{e=1}^E$, which are subsequently passed to the transformer decoder for final segmentation.
\subsection{Transformer Decoder}
Finally, the transformer decoder~\cite{cheng2022masked} iteratively refines the initially aligned textual queries with $z_{out}$ through $N$ masked attention layers and outputs refined queries $\mathbf{Q}^{(N)}$. These are projected into mask embeddings $\mathbf{M_E}$ to generate pixel-level mask with the per-pixel embeddings $\mathbf{Z}_{\text{pixel}}$, expressed as $\hat{\mathbf{Y}}_{\text{mask}} = \mathbf{M_E} \cdot \mathbf{Z}_{\text{pixel}}$. Each query embedding from $\mathbf{Q}^{(N)}$ is classified through a linear projection to produce class logits $\hat{\mathbf{C}}$. The final segmentation prediction is obtained by combining the pixel-level mask and the class scores as $\hat{\mathbf{Y}} = \hat{\mathbf{C}}^\top \cdot \hat{\mathbf{Y}}_{\text{mask}} \quad $, where $\hat{\mathbf{C}}^\top $ denotes transpose of $\hat{C}$. 

\subsection{Cross-View and Modal Consistency Regularisation (CMCR)}
Limited labeled data restricts the model's exposure to diverse visual conditions, causing overfitting to the annotated subset and weakening its ability to generalize across varying image distributions. In mask-transformer architectures, this insufficient supervision can amplify early misalignments between pixel features and textual queries during iterative decoding~\cite{pak2024textual}, leading to cumulative errors and reduced domain robustness.

To overcome this, we designed Cross-View and Modal Consistency Regularization (CMCR), which extracts supervisory signals from unlabeled data and reinforces domain robustness. CMCR enforces prediction consistency across multiple augmented views of the same image, compelling the model to maintain stable vision–language alignment under appearance variations. Applied at every decoder layer, it prevents error accumulation and enables textual queries to adapt effectively to diverse image contexts under limited supervision, thus enhancing the model's semantic understanding. We adopt a three-view augmentation strategy comprising the original image $I$, a weakly augmented version $I_w$ (e.g., Gaussian blur, slight brightness or contrast changes), and a strongly augmented version $I_s$ (e.g., Cutout, color jitter, and geometric distortions) following UniMatch~\cite{Yang_2023_CVPR}. The original and weak views preserve intrinsic domain characteristics—such as color distribution, structure, and contextual layout—while the strong view introduces appearance diversity. CMCR enforces consistency across all three views, i.e., original–weak and original–strong pairs, allowing the model to anchor on domain-stable cues while learning invariance to superficial perturbations. This balance ensures domain robustness without eroding domain awareness.

Given an unlabeled image $I$ and its augmented counterparts $(I_w, I_s)$, the transformer decoder at each stage $n \in \{1, \dots, N\}$ produces a mask prediction $(\hat{\mathbf{Y}}^{n}, \hat{\mathbf{Y}}_{w}^{n}, \hat{\mathbf{Y}}_{s}^{n})$, class predictions by each query $i \in \{1, \dots, K.E\}$ :  $(\hat{\mathbf{C}}_{i}^{n}, \hat{\mathbf{C}}_{w,i}^{n}, \hat{\mathbf{C}}_{s,i}^{n})$, and pixel-text attention maps $(\mathbf{A}_{i}^{n}, \mathbf{A}_{w,i}^{n}, \mathbf{A}_{s,i}^{n})$. The CMCR objective enforces prediction and pixel-text consistency across these augmented views for each query:
\vspace{-0.7\baselineskip}
\begin{equation}
\footnotesize
\ell_{\text{CMCR}} = \sum_{n=1}^{N} [  \, \ell_{\text{mask}}(\hat{\mathbf{Y}}^{n}, \hat{\mathbf{Y}}_{w}^{n}, \hat{\mathbf{Y}}_{s}^{n})+ \sum_{i=1}^{K} (  \, \ell_{\text{class}}(\hat{\mathbf{C}}_{i}^{n}, \hat{\mathbf{C}}_{w,i}^{n}, \hat{\mathbf{C}}_{s,i}^{n})+  \, \ell_{\text{align}}(\mathbf{A}_{i}^{n}, \mathbf{A}_{w,i}^{n}, \mathbf{A}_{s,i}^{n}))]. 
\label{cmcr_eq}
\end{equation}

\noindent Here, $\ell_{\text{mask}}$ enforces global segmentation consistency across augmented views, while $\ell_{\text{class}}$ and $\ell_{\text{align}}$ regularize query-specific class predictions and pixel–text correspondences. The details of these losses are provided in the supplementary. Our overall training objective combines segmentation( $\ell_{seg}$ ), and regularization ( $\ell_{reg}$ ) from TQDM\cite{pak2024textual}, along with diversity and CMCR regularization, expressed as
\vspace{-0.7\baselineskip}
\begin{equation}
\ell_{\text{total}} = \ell_{\text{seg}} + \lambda_{\text{reg}} \ell_{\text{reg}} + \lambda_{\text{div}} \ell_{\text{div}} + \lambda_{\text{cmcr}} \ell_{\text{cmcr}},
\end{equation}
where $\{\lambda_{\text{reg}}, \lambda_{\text{div}}, \lambda_{\text{cmcr}}\}$ are weighting factors.

\begin{table}[t]
\centering
\tiny
\caption{Comparison of HVLFormer with State-Of-The-Art (SOTA) SSS methods on Pascal VOC. The mIoU (\%) is reported across varying labeled splits. The best results are shown in {\color{red}red}, the previous SOTA in {\color{blue}\underline{blue}}, while relative improvements are indicated in {\color{ForestGreen}\underline{green}} and {\color{brown}\underline{brown}}, and \textbf{bold} denotes variants of our method.. $^\dagger$: reproduced by \cite{hoyer2024semivl}.}
\label{tab1}
\begin{tabular}{l c c c c c c l}
\toprule
\textbf{Methods} & \textbf{Backbones} & \textbf{1/115} & \textbf{1/58} & \textbf{1/29} & \textbf{1/14} & \textbf{1/7}  &\#Params\\
 & & \textbf{92} & \textbf{183} & \textbf{366} & \textbf{732} & \textbf{1464}  &\\
\midrule
PseudoSeg~\cite{ijcai2021pseudoseg} & R101 & 57.6 & 65.5 & 69.1 & 72.4 & --  &59.5\\
CPS~\cite{chen2021cps} & R101 & 64.1 & 67.4 & 71.7 & 75.9 & --  &59.5\\
ST++~\cite{yang2022st++} & R101 & 65.2 & 71.0 & 74.6 & 77.3 & 79.1  &59.5\\
U$^2$PL~\cite{wang2022semi} & R101 & 68.0 & 69.2 & 73.7 & 76.2 & 79.5  &59.5\\
PCR~\cite{xu2022semi} & R101 & 70.1 & 74.7 & 77.2 & 78.5 & 80.7  &59.5\\
ESL~\cite{guo2023esl} & R101 & 71.0 & 74.0 & 78.1 & 79.5 & 81.8  &59.5\\
LogicDiag~\cite{liang2023logic} & R101 & 73.3 & 76.7 & 77.9 & 79.4 & --  &59.5\\
UniMatch V1~\cite{Yang_2023_CVPR} & R101 & 75.2 & 77.2 & 78.9 & 79.9 & 81.2  &59.5\\
3-CPS~\cite{he2023three} & R101 & 75.7 & 77.7 & 80.1 & 80.9 & 82.0  &59.5\\
\midrule
ZegCLIP$^\dagger$~\cite{Zhou2023} & ViT-B & 69.3 & 74.2 & 78.7 & 81.0 & 82.0  &88.0\\
ZegCLIP+Unimatch V1$^\dagger$ & ViT-B & 78.0 & 80.3 & 80.9 & 82.8 & 83.6  &88.0\\
UniMatch V1$^\dagger$~\cite{Yang_2023_CVPR} & ViT-B & 77.9 & 80.1 & 82.0 & 83.3 & 84.0  &88.0\\
Pseudo-SD~\cite{zhao2025pseudo}& ViT-B & 78.9 & 79.2 & 80.7 & 81.9 & 82.7  &88.0\\
SemiVL~\cite{hoyer2024semivl} & ViT-B & 84.0 & 85.6 & 86.0 & 86.7 & 87.3  &88.0\\
 TQDM (baseline)~\cite{pak2024textual} & ViT-B 
& 70.4& 74.1& 77.9& 82.3&86.7 &105.0\\
\textbf{Ours} & 
ViT-B 
& \textbf{87.3}& \textbf{88.8}& \textbf{89.4}& \textbf{90.0}& \textbf{90.2} &107.0\\
 \textbf{Ours} & EVA02-S& \textbf{84.8}& \textbf{86.2}& \textbf{87.1}& \textbf{88.5}& \textbf{89.3}&\textbf{40.0}\\
 UniMatch V2~\cite{10839097} & DINOv2-S & 79.0 & 85.5 & 85.9 & 86.7 &87.8  &24.8\\
POS~\cite{sun2025two} & DINOv2-S & 80.7 & 86.8 & 87.2 & 87.5 & 88.3  &24.8\\
 UniMatch V2~\cite{10839097} & DINOv2-B& \underline{\textcolor{blue}{86.3}} & \underline{\textcolor{blue}{87.9}} & \underline{\textcolor{blue}{88.9}} & \underline{\textcolor{blue}{90.0}} & \underline{\textcolor{blue}{90.8}}  &98.0\\
TQDM (baseline)~\cite{pak2024textual} & SigLIP2& 76.1 & 79.5 & 82.5 & 86.1 & 88.6  &107.0\\
\textbf{Ours} & SigLIP2& \textbf{\textcolor{red}{89.4}} & \textbf{\textcolor{red}{90.1}} & \textbf{\textcolor{red}{90.9}} & \textbf{\textcolor{red}{91.5}} & \textbf{\textcolor{red}{91.8}}  &109.0\\
\midrule
Gain (SOTA)& & \textcolor{ForestGreen}{+3.1} & \textcolor{ForestGreen}{+2.2} & \textcolor{ForestGreen}{+2.0} & \textcolor{ForestGreen}{+1.5} & \textcolor{ForestGreen}{+1.0}  &\\
Gain (baseline) & &\textcolor{brown}{+13.3} & \textcolor{brown}{+10.6}& \textcolor{brown}{+8.4} & \textcolor{brown}{+5.4} & \textcolor{brown}{+3.2}  &\\
\bottomrule
\end{tabular}
\end{table}

\section{Experiments}
\subsection{Implementation Details}

\textbf{Network Architecture}. Our model's vision encoder $E_I$~\cite{dosovitskiy2020vit} and text encoder $E_T$~\cite{vaswani2017attention} are initialized using one of three backbones: CLIP~\cite{Radford_2021_ICML}, SigLip2 \cite{tschannen2025siglip}, or an EVA02-CLIP~\cite{fang2023eva} with a patch size of 16 or 14, respectively. We employ an 8-token learnable prompt $p_k$ to generate attribute-rich textual descriptions per dataset-class. The hierarchical queries in HTQG are generated with  $E = 3$ MLP heads; hence, three queries per class, strictly aligned with three feature maps from pixel-decoder. The multi-scale pixel decoder is adopted from DN-DETR~\cite{li2022dn}, with $M=6$ layers. The transformer decoder follows Mask2Former~\cite{cheng2022masked} with $N = 9$ masked attention layers. We adopt TQDM\cite{pak2024textual} as the baseline. We utilize mean Intersection over Union (mIoU) as metric, while results are averaged over three random data sampling seeds.

\begin{figure}[!tbp]
  \centering
  \begin{tabular}{c@{ }c@{ }c@{ }c|cccc}
     Image&UniM. V2& Ours& GT &Image&UniM. V2&Ours&GT 
\\
     \includegraphics[width=.115\textwidth]{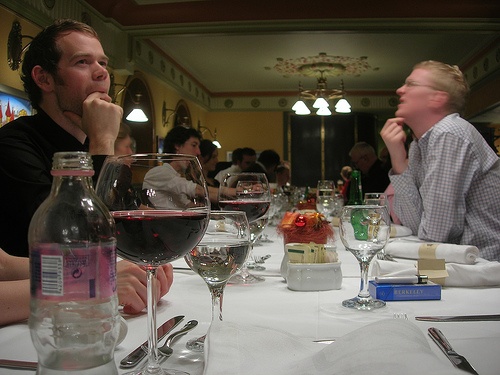}&
    \includegraphics[width=.115\textwidth]{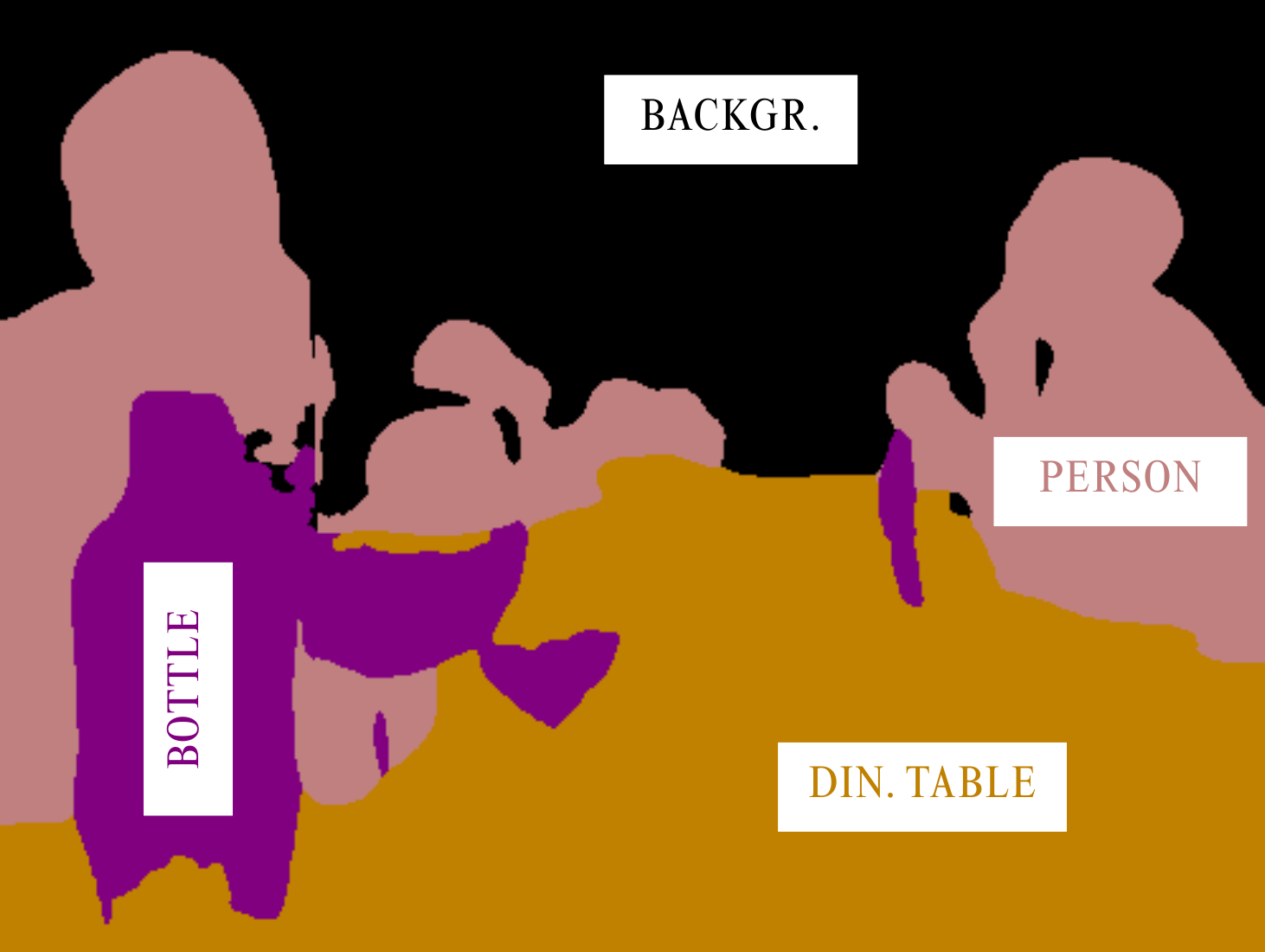}&
    \includegraphics[width=.115\textwidth]{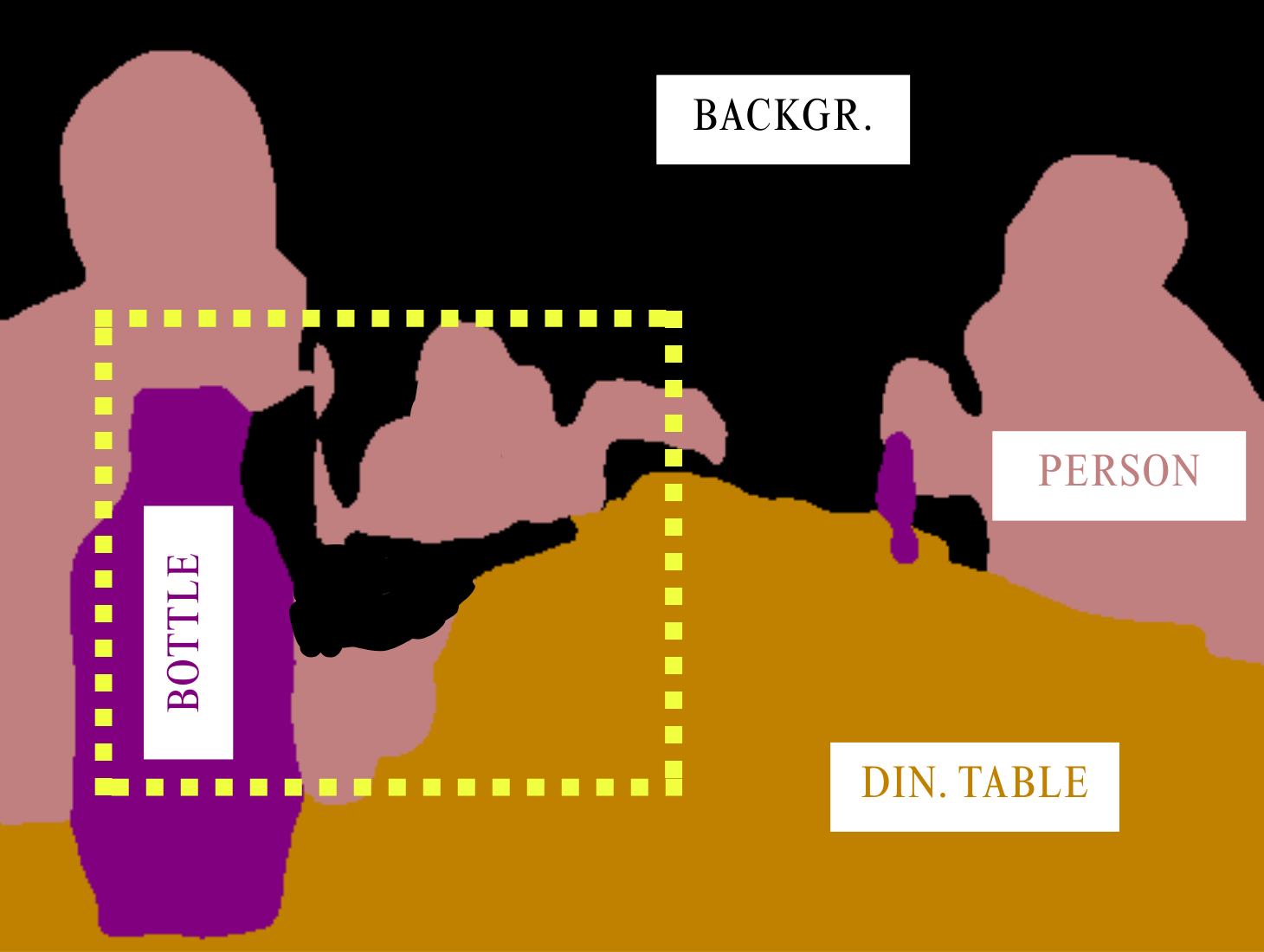}&
    \includegraphics[width=.115\textwidth]{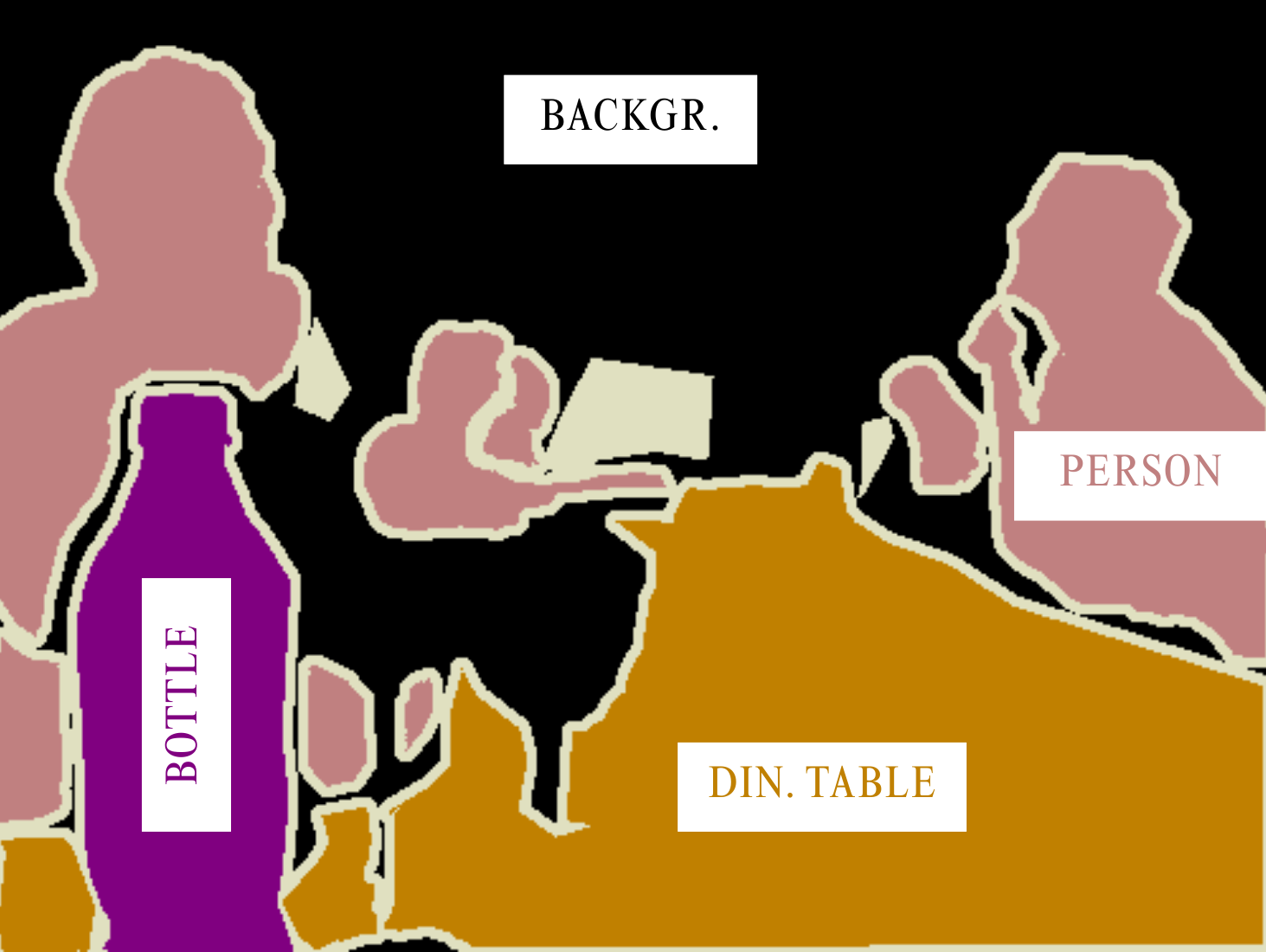} & \includegraphics[width=.115\textwidth]{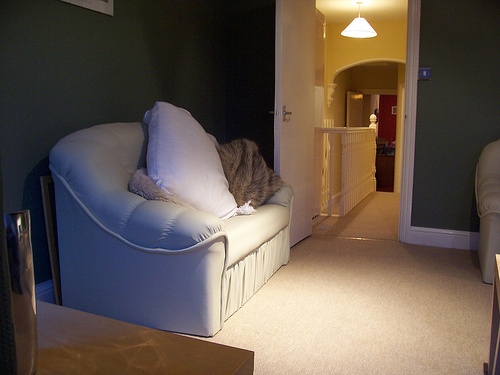}& \includegraphics[width=.115\textwidth]{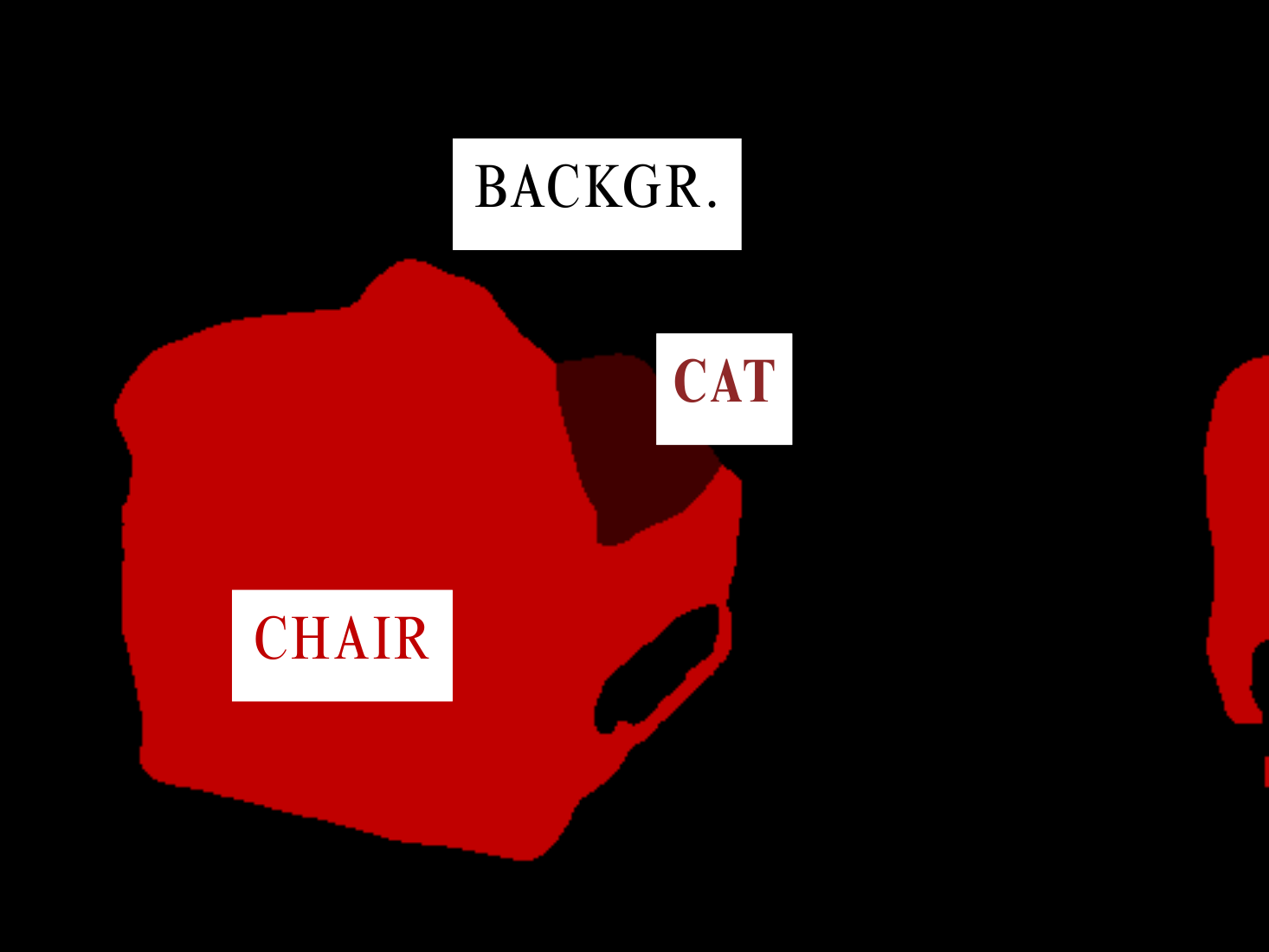}& \includegraphics[width=.115\textwidth]{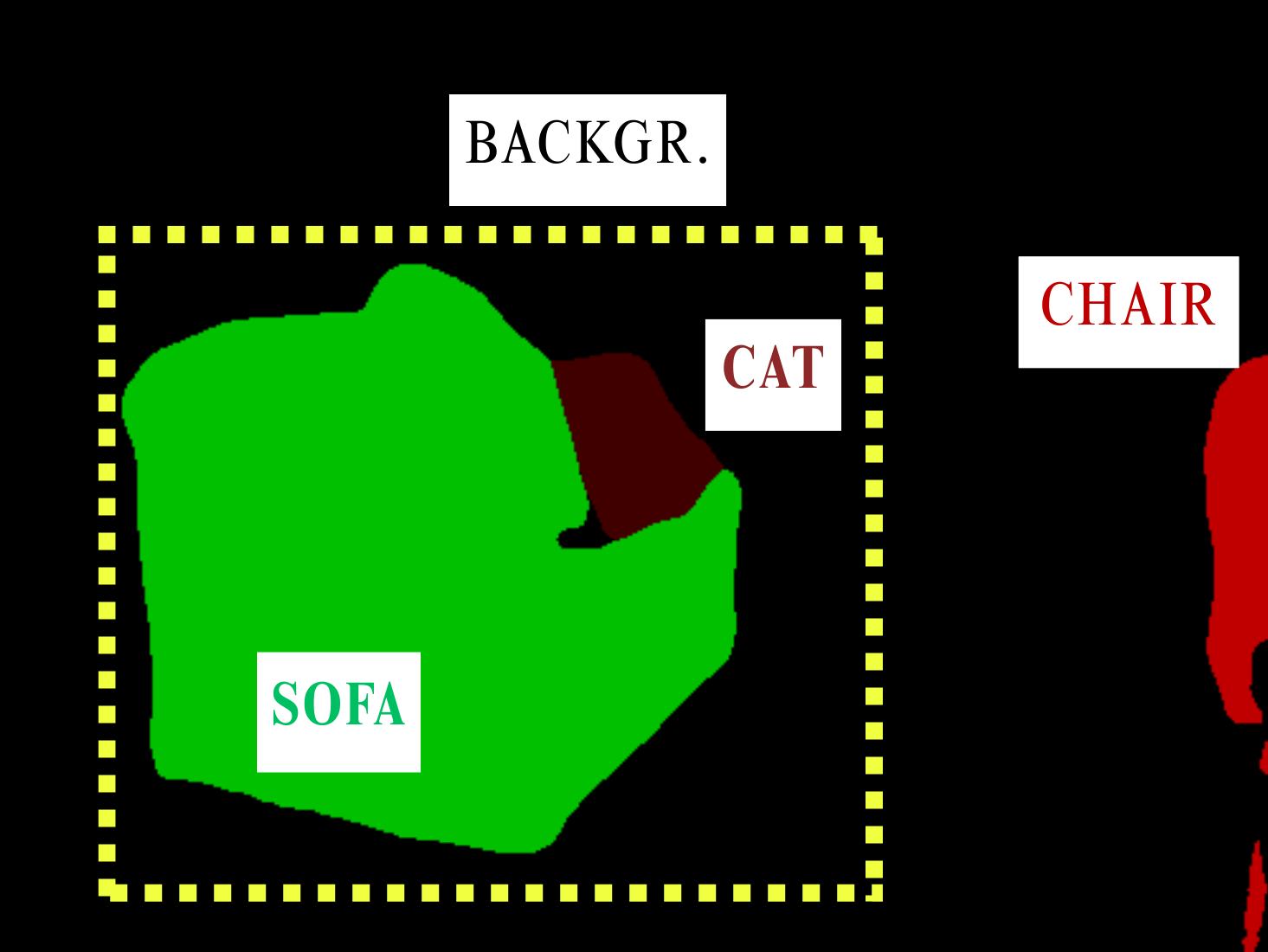}&\includegraphics[width=.115\textwidth]{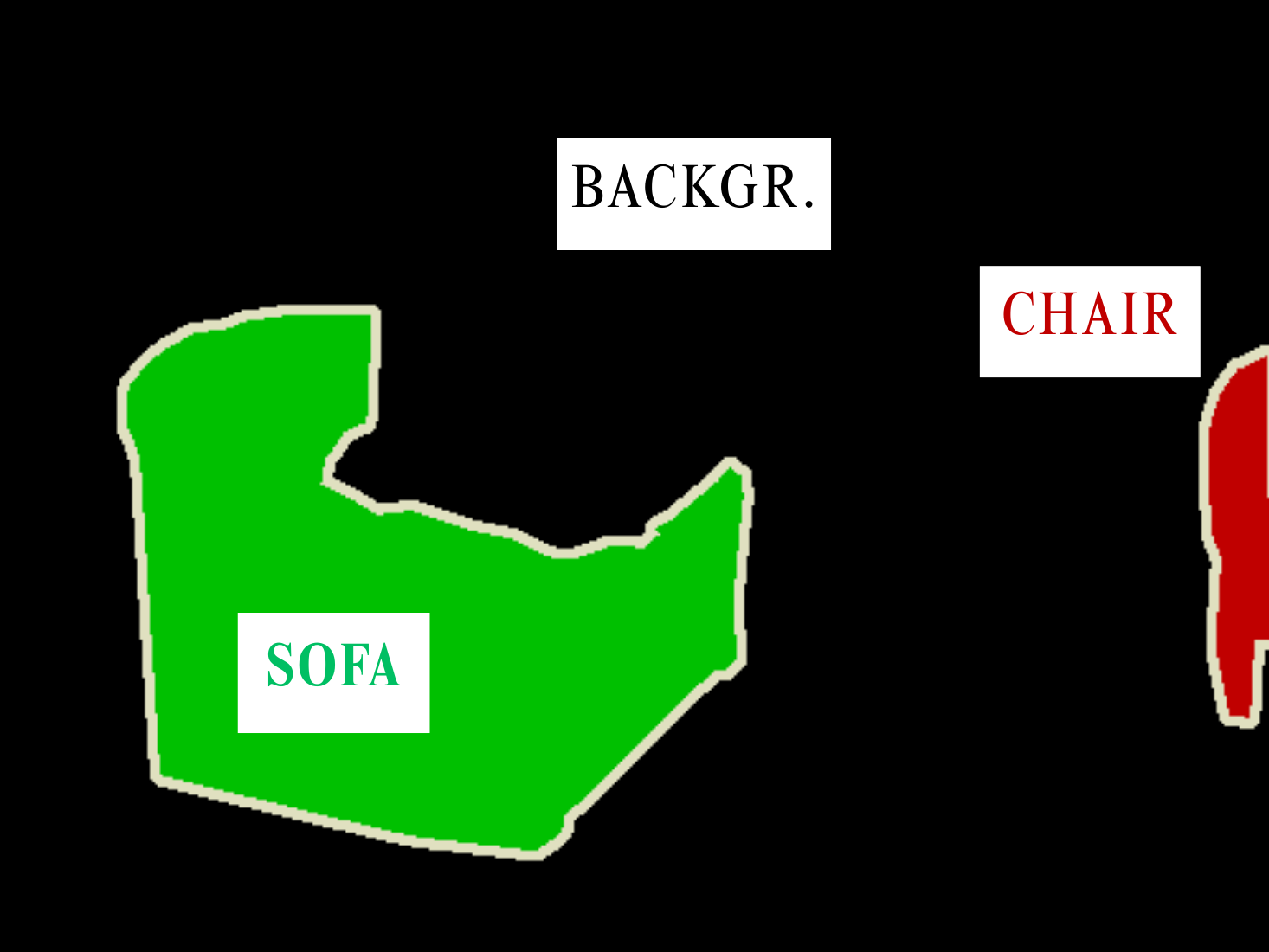}
\\
     \includegraphics[width=.115\textwidth]{images/coco1.jpg} &
    \includegraphics[width=.115\textwidth]{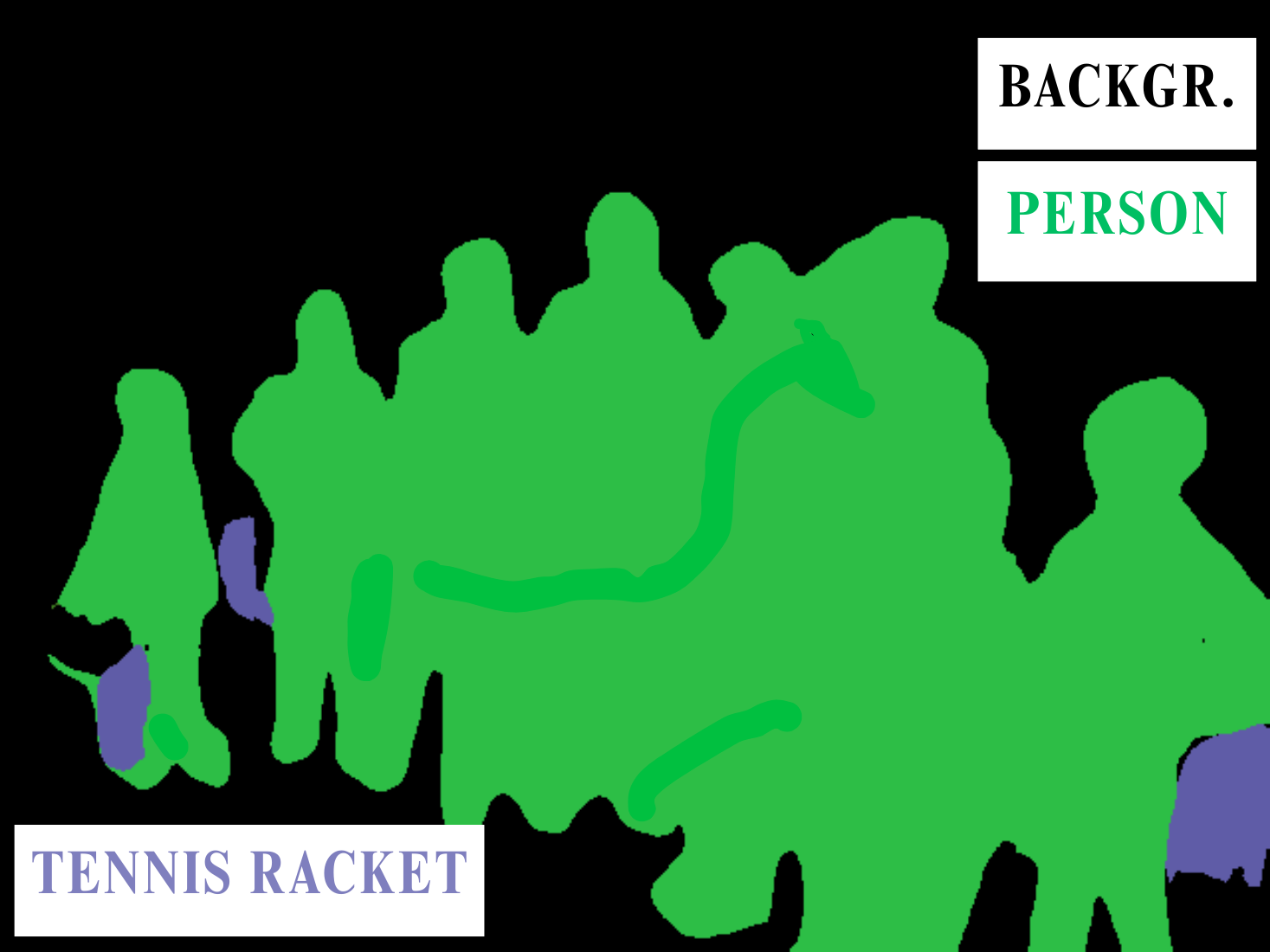} &
    \includegraphics[width=.115\textwidth]{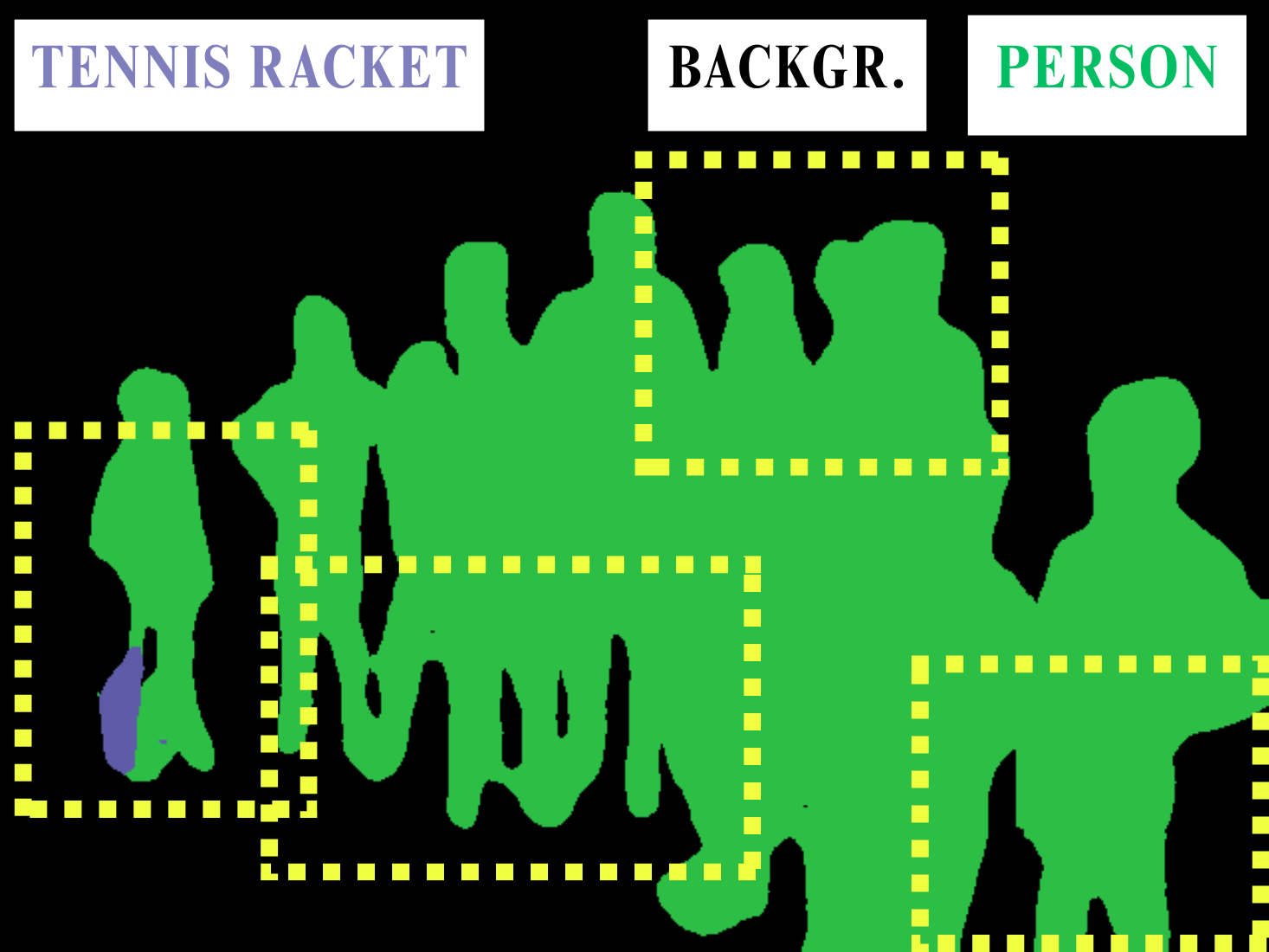} &
    \includegraphics[width=.115\textwidth]{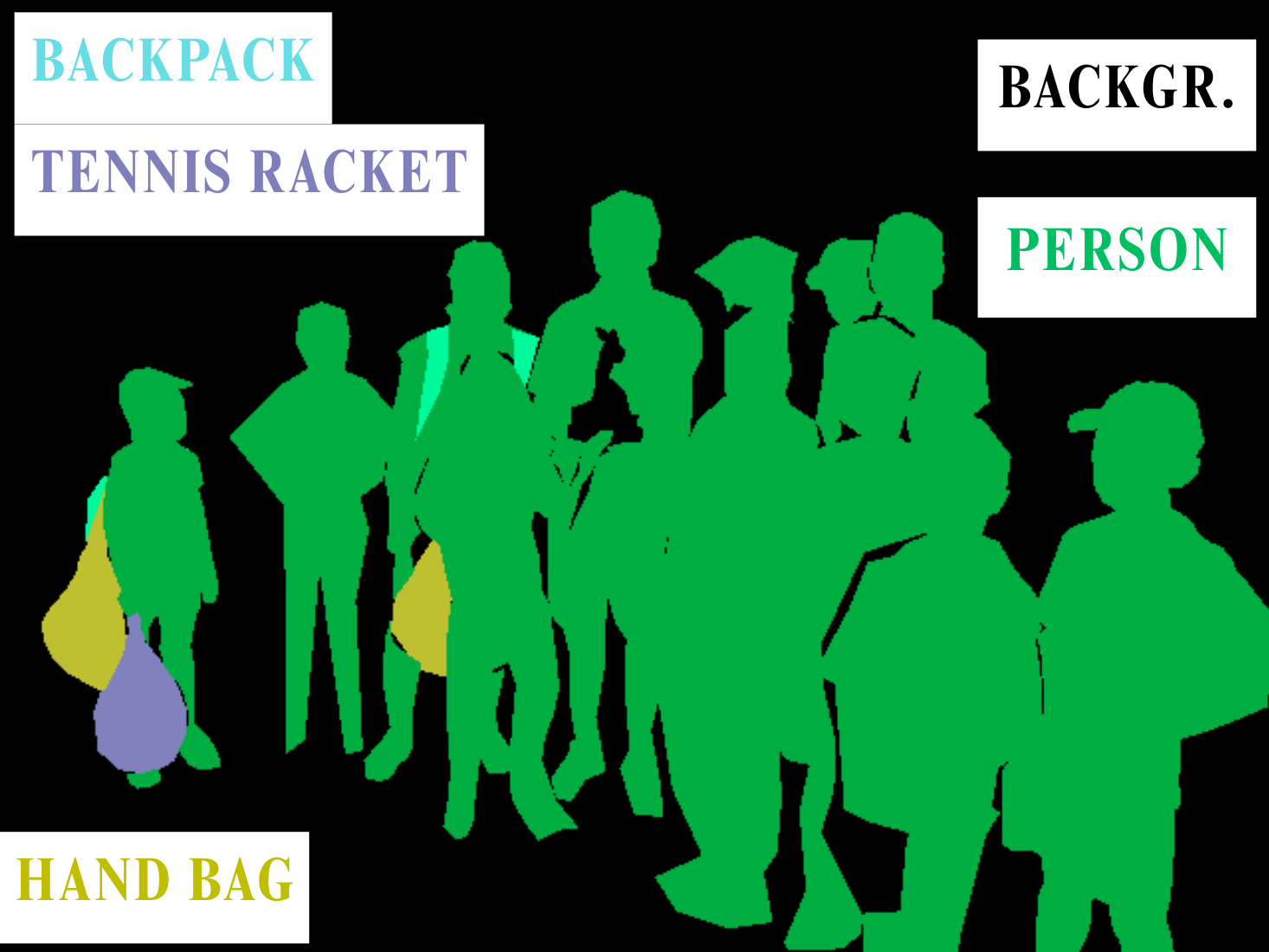}   & \includegraphics[width=.115\textwidth]{images/coco5.jpg} & \includegraphics[width=.115\textwidth]{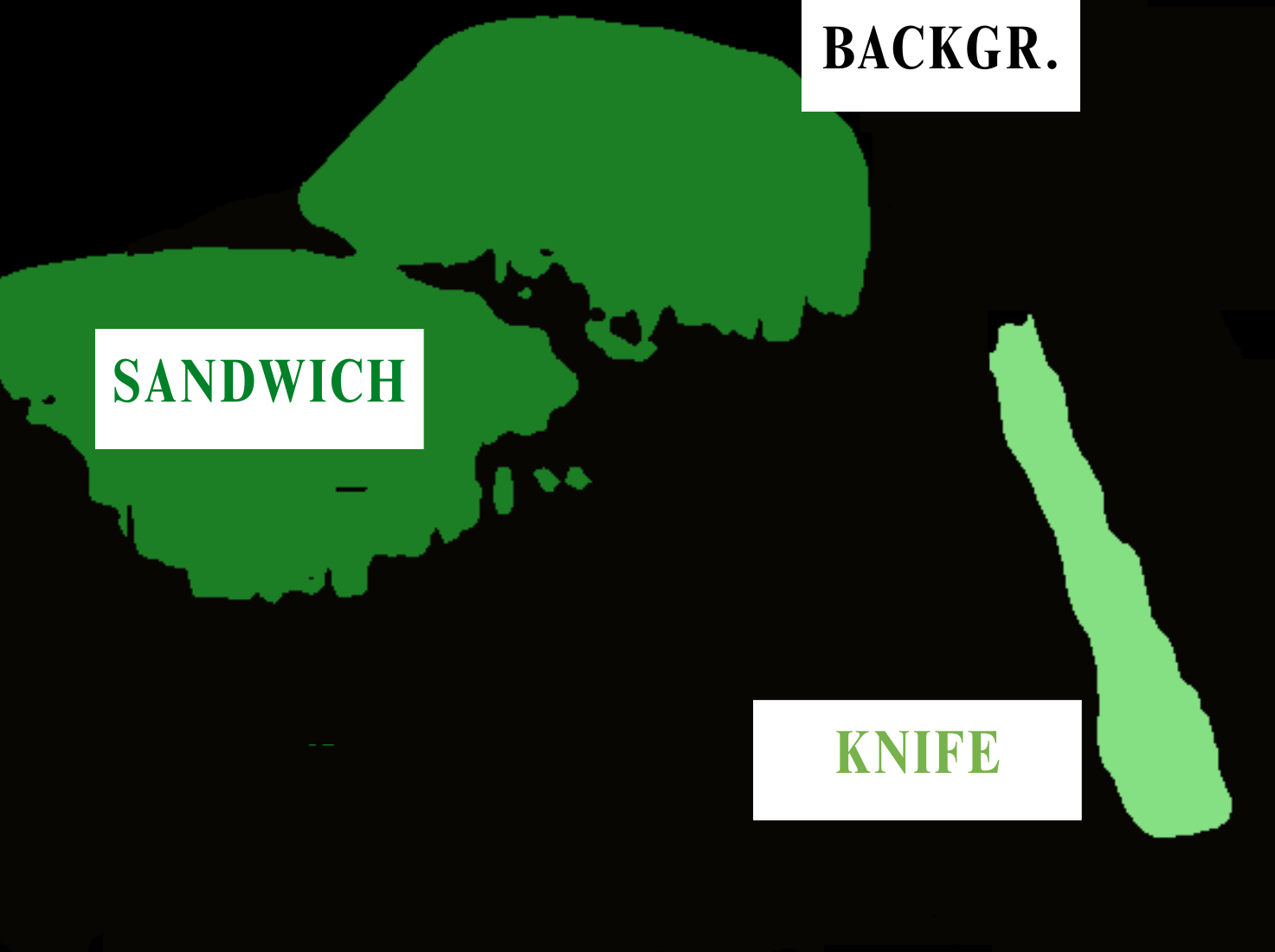} & \includegraphics[width=.115\textwidth]{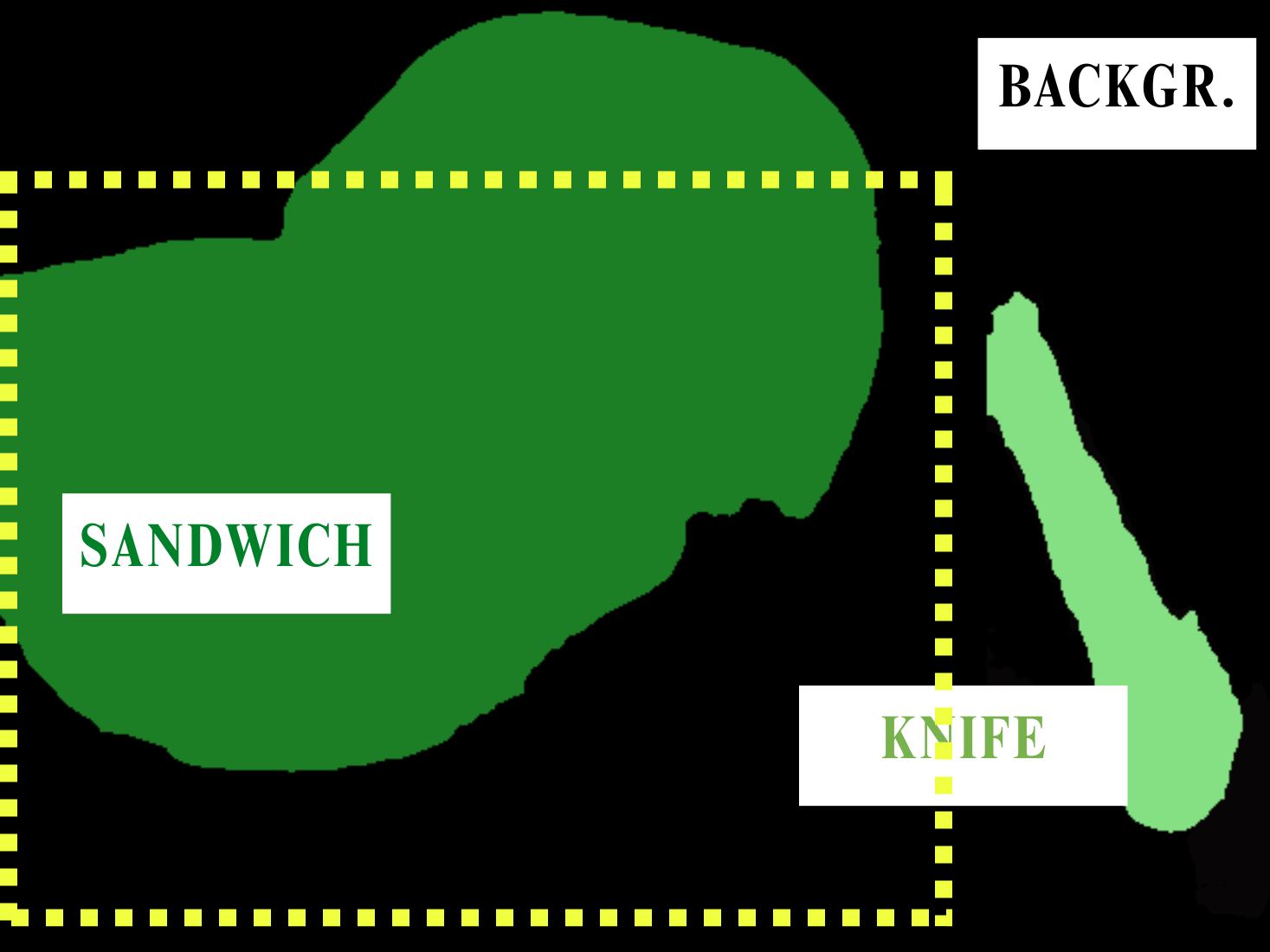} &\includegraphics[width=.115\textwidth]{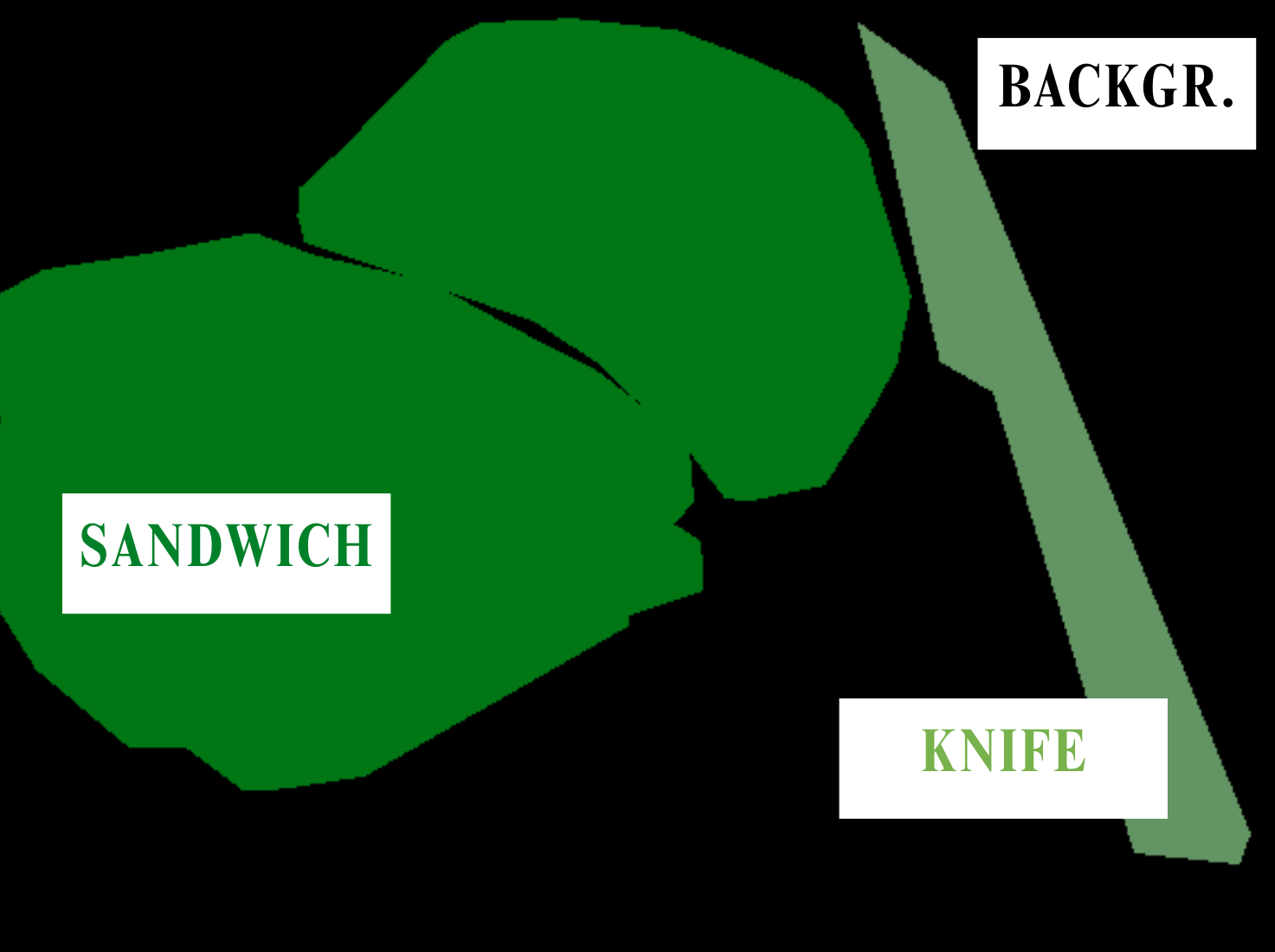} \\
  \end{tabular}
  \caption{The Pascal VOC with 92 labels (top) and COCO with 232 labels (bottom). HVLFormer correctly disambiguates visually similar classes, suppresses over-segmentation in cluttered regions, and accurately identifies multiple persons in dense crowds, demonstrating enhanced semantic understanding in low-label regime.}
  \label{fig:visual_comp}
\vspace{-4mm}
\end{figure}

\noindent\textbf{Training}. We benchmark on Pascal VOC~\cite{pascal_voc}, COCO~\cite{coco}, ADE20K~\cite{ade20k}, and Cityscapes~\cite{cityscapes} using two 24 GB NVIDIA GeForce RTX-3090 GPUs. Following SemiVL~\cite{hoyer2024semivl}, each iteration processes a mixed batch of eight labeled and eight unlabeled images. Inputs are randomly cropped to $512 \times 512$ pixels, except for Cityscapes, where we use $801 \times 801$. The model is trained for 80, 10, 40, and 240 epochs on VOC, COCO, ADE20K, and Cityscapes, respectively, using AdamW~\cite{Ding2022}. The initial learning rates are set to $10^{-4}$ with a polynomial decay of 0.9. Linear warm-up~\cite{10.1007/978-3-031-91835-3_19} is applied for $t_{\text{warm}} = 1.5\text{k}$ iterations. The weighting factors for each objective are set to $\{\lambda_{\text{reg}}, \lambda_{\text{div}}, \lambda_{\text{cmcr}}\} = \{1, 1, 5\}$, with a detailed hyperparameter study provided in the supplementary material. Image augmentation in the CMCR module follows Unimatch V2~\cite{10839097}, with weak augmentations that apply mild photometric changes, and strong augmentations that include color jitter, CutMix, and geometric distortions. Following~\cite{Nadeem_2025_CVPR}, rare-class sampling is applied.

\subsection{\textbf{Comparison with the State-Of-The-Art Methods}}

We benchmark HVLFormer against leading SSS methods across four benchmarks. The model achieves State-Of-The-Art (SOTA) performance with consistent gains over both VLM and non-VLM methods. By leveraging domain-aware and robust language semantics for segmentation mask generation, HVLFormer enhances model semantic understanding under low-label conditions, demonstrating larger improvements at lower labeled training ratios and strong label efficiency overall. Even with the shallower EVA02-S ~\cite{fang2023eva} backbone, HVLFormer (40M) outperforms all VLM-based SOTA, including SemiVL (88M) \cite{hoyer2024semivl} and Pseudo-SD (88M) \cite{zhao2025pseudo}, highlighting its effectiveness in utilizing global textual priors for enhanced semantic understanding in low-label regimes.
 As shown in Fig.~\ref{fig:visual_comp}, HVLFormer produces precise segmentations, effectively distinguishing confusing or rare class pairs such as sofa–chair, and avoiding background misclassifications like racket nets, demonstrating strong domain awareness, contextual reasoning, and semantic consistency.

\noindent\textbf{Comparisons on Pascal VOC.} We vary the number of labeled images from 92 to 1,464 on the Pascal VOC dataset (10,582 training images; 1,464 labeled). Tab.~\ref{tab1} summarizes results across these splits, showing that HVLFormer consistently outperforms prior VLM-based methods, including Pseudo-SD~\cite{zhao2025pseudo} and SemiVL~\cite{hoyer2024semivl}, with gains of up to +5.4 mIoU under the lowest labeled split. Compared with its baseline TQDM \cite{pak2024textual}, our improvements range from +13.3 mIoU to +3.2 mIoU, confirming that conditioning domain-invariant queries with domain awareness leads to substantial benefits in low-annotation regimes. With only 14\% labeled data, it achieves a new SOTA of 91.8 mIoU over Unimatch V2's~\cite{10839097} 90.8\%. 

\noindent\textbf{Comparison on COCO.} The COCO dataset (118k training images; 81 classes) introduces a higher degree of complexity for SSS due to its category diversity and fine-grained inter-class overlap. Despite this, HVLFormer achieves SOTA, with a +9.3 mIoU gain using only 232 labeled images and up to a +19.2 mIoU improvement over its baseline depicted in Tab \ref{tab:coco}. Its consistent gains across splits indicate reduced ambiguity in segmenting visually similar and rare classes, alongside improved modeling of intra-class variation. Extended performance comparisons are provided in supp.

\noindent\textbf{Comparison on ADE20K.} We present an even broader scene-parsing challenge using ADE20K (20.21k train images; 150 classes). Tab.~\ref{tab:ade} shows that HVLFormer achieves a +13.5 mIoU gain over its baselines with only 158 labels, while achieving a +4.8 mIoU over existing SOTA with 316 training labels, further validating its effectiveness across several classes and challenging datasets.

\noindent\textbf{Comparison on Cityscapes.} On Cityscapes (2.975k training images; 19 classes), HVLFormer outperforms SemiVL~\cite{hoyer2024semivl} by 3.4\% with 100 labels and surpasses UniMatch V2 \cite{10839097} by +0.6 mIoU with 186 labels, as shown in Tab.~\ref{tab:city}. While the overall gains are smaller compared to other benchmarks, this is due to the presence of numerous small and distant object classes (e.g., poles, traffic signs) that are not part of the captions in VLM pre-training.
\begin{table}[t]
\centering
\tiny

\begin{minipage}[t]{0.49\columnwidth}
\centering
\caption{HVLFormer comparison on COCO under SSS.}
\label{tab:coco}

\begin{tabular}{l c c c c c c}
\toprule
Method & BB & 1/512 & 1/256 & 1/128 & 1/64 & 1/32 \\
 & & (232) & (463) & (925) & (1.8k) & (3.7k) \\
\midrule
UniM V1~\cite{Yang_2023_CVPR}& XC65 & 31.9 & 38.9 & 44.4 & 48.2 & 49.8 \\
LogicD.~\cite{liang2023logic} & XC65 & 33.1 & 40.3 & 45.4 & 48.8 & 50.5 \\
UniM V1$^\dagger$~\cite{Yang_2023_CVPR}& ViT-B & 36.6 & 44.1 & 49.1 & 53.5 & 55.0 \\
Pseudo-SD~\cite{zhao2025pseudo} & ViT-B & 39.9 & 44.8 & 48.2 & 50.2 & 51.9 \\
SemiVL~\cite{hoyer2024semivl} & ViT-B & 50.1 & 52.8 & 53.6 & 55.4 & 56.5 \\

\textbf{Ours} & ViT-B & \textbf{54.9} & \textbf{56.7} & \textbf{61.2} & \textbf{62.9} & \textbf{65.1} \\
\textbf{Ours} & E.02-S& \textbf{52.7} & \textbf{54.8} & \textbf{56.2} & \textbf{58.5} & \textbf{61.1} \\

POS~\cite{sun2025two} & D.v2-S& 40.9 & 46.8 & 54.3 & 56.2 & 57.7 \\
UniM V2~\cite{10839097}& D.v2-B& \underline{\textcolor{blue}{47.9}} & \underline{\textcolor{blue}{55.8}} & \underline{\textcolor{blue}{58.7}} & \underline{\textcolor{blue}{60.4}} & \underline{\textcolor{blue}{63.3}} \\

TQDM~\cite{pak2024textual} & SigLIP2 & 40.2 & 48.7 & 54.3 & 58.9 & 62.7 \\
\textbf{Ours} & SigLIP2 & \textbf{\textcolor{red}{59.4}} & \textbf{\textcolor{red}{61.3}} & \textbf{\textcolor{red}{63.7}} & \textbf{\textcolor{red}{65.1}} & \textbf{\textcolor{red}{67.4}} \\
\bottomrule
\end{tabular}

\end{minipage}
\hfill
\begin{minipage}[t]{0.49\columnwidth}
\centering
\caption{HVLFormer comparison on Cityscapes under SSS.}
\label{tab:city}

\begin{tabular}{l c c c c c c}
\toprule
Method & BB & 1/30 & 1/16 & 1/8 & 1/4 & 1/2 \\
 & & (100) & (186) & (372) & (744) & (1.4k) \\
\midrule
P.Seg~\cite{ijcai2021pseudoseg} & R101 & 61.0 & -- & 69.8 & 72.4 & -- \\

UniM V1$^\dagger$~\cite{Yang_2023_CVPR}& ViT-B & 73.8 & 76.6 & 78.2 & 79.1 & 79.6 \\
Pseudo-SD~\cite{zhao2025pseudo} & ViT-B & 76.1 & 78.5 & 79.5 & 80.5 & 81.5 \\
SemiVL~\cite{hoyer2024semivl} & ViT-B & \underline{\textcolor{blue}{76.2}} & 77.9 & 79.4 & 80.3 & 80.6 \\

TQDM~\cite{pak2024textual} & ViT-B & 65.2 & 72.8 & 77.1 & 80.7 & 81.8 \\

\textbf{Ours} & ViT-B & \textbf{78.1} & \textbf{83.0} & \textbf{84.6} & \textbf{84.8} &\textbf{85.1} \\
\textbf{Ours} & E.02-S& \textbf{77.5} & \textbf{81.3} &\textbf{ 83.7} & \textbf{84.0} & \textbf{84.3} \\

POS~\cite{sun2025two} & D.v2-S& 40.9 & 81.4 & 82.7 & 83.0 & 83.2 \\
UniMV2~\cite{10839097}& D.v2-B& -- & \underline{\textcolor{blue}{83.6}} & \underline{\textcolor{blue}{84.3}} & \underline{\textcolor{blue}{84.5}} & \underline{\textcolor{blue}{85.1}} \\

TQDM~\cite{pak2024textual} & SigLIP2 & 69.1 & 75.5 & 79.2 & 81.6 & 83.9 \\
\textbf{Ours} & SigLIP2 & \textbf{\textcolor{red}{79.6}} & \textbf{\textcolor{red}{84.1}} & \textbf{\textcolor{red}{84.9}} & \textbf{\textcolor{red}{85.1}} & \textbf{\textcolor{red}{85.3}} \\
\bottomrule
\end{tabular}

\end{minipage}

\end{table}

\begin{table}[t]
\centering
\tiny

\begin{minipage}[t]{0.49\columnwidth}
\centering
\caption{HVLFormer comparison on ADE20K under SSS.}
\label{tab:ade}

\begin{tabular}{l c c c c c c}
\toprule
Method & BB & 1/512 & 1/256 & 1/128 & 1/64 & 1/32 \\
 & & (232) & (463) & (925) & (1.8k) & (3.7k) \\
\midrule
AEL~\cite{Hu2021Semi} & R101 & -- & -- & 28.4 & 33.2 & 38.0 
\\
UniMatch V1$^\dagger$~\cite{Yang_2023_CVPR} & ViT-B & 18.4 & 25.3 & 31.2 & 34.4 & 38.0 
\\
Pseudo-SD~\cite{zhao2025pseudo} & ViT-B & -- & 28.1 & 33.3 & 35.2 & 39.3 
\\
SemiVL~\cite{hoyer2024semivl} & ViT-B & \underline{\textcolor{blue}{28.1}} & 33.7 & 35.1 & 37.2 & \underline{\textcolor{blue}{39.4}} 
\\
\textbf{Ours} & ViT-B & \textbf{35.2}& \textbf{39.7}& \textbf{46.6}& \textbf{47.5}& \textbf{51.0}
\\

\textbf{Ours} & EVA02-S& \textbf{33.8}& \textbf{35.3}& \textbf{36.9}& \textbf{39.1}& \textbf{45.7}
\\

UniM V2~\cite{10839097}& D.v2-B& -- & \underline{\textcolor{blue}{38.7}} & \underline{\textcolor{blue}{45.0}} & \underline{\textcolor{blue}{46.7}} & \underline{\textcolor{blue}{49.8}} 
\\
TQDM~\cite{pak2024textual}& SigLIP2& 25.3 & 31.8 & 39.9 & 43.2 & 48.1 
\\

 \textbf{Ours} & SigLIP2& \textbf{\textcolor{red}{41.6}} & \textbf{\textcolor{red}{43.5}} & \textbf{\textcolor{red}{48.2}} & \textbf{\textcolor{red}{49.8}} &\textbf{\textcolor{red}{52.6}} 
\\\hline
\end{tabular}

\end{minipage}
\hfill
\begin{minipage}[t]{0.435\columnwidth}
\centering
\scriptsize
\caption{Study on Semantic Relevance Estimator (SRE) with different variants on Pascal VOC and COCO.}
\label{tab:sre}

\begin{tabular}{l |c |c }
\toprule
Configuration& Pascal& COCO
\\
 & $mIoU_{92}$ & $mIoU_{232}$ 
\\
\midrule
No SRE
& 88.3& 57.8
\\

Fixed threshold & 88.7& 58.1
\\
 ( $s_k>0.5$)& &\\
Soft weighing & \textcolor{red}{89.4}& \textcolor{red}{59.4}\\

 (weigh queries by $s_k$)& &\\
 \hline
\end{tabular}

\end{minipage}

\end{table}

 \begin{table}[t]
\centering
\begin{minipage}[t] {0.49\columnwidth}
\centering
\caption{Ablation on Pascal VOC and COCO under SSS. Low-label splits benefit more than higher-label splits.}
\resizebox{\columnwidth}{!}{
\begin{tabular}{ll|cc|ll}
\toprule
\multicolumn{2}{l|}{Configuration} & \multicolumn{2}{c|}{Pascal} & \multicolumn{2}{c}{COCO}\\ \cline{3-6}
\multicolumn{2}{c|}{} & $mIoU_{92}$ & $mIoU_{1464}$ & $mIoU_{232}$ & $mIoU_{3700}$\\
\midrule
\multicolumn{2}{l|}{Baseline} & 76.1 & 88.6 & 40.2 & 62.7\\
\multicolumn{2}{l|}{+HTQG (Dataset attr)} & 77.3 & 88.9 & 41.9 & 63.2\\
\multicolumn{2}{l|}{+HTQG (Dataset attr + HQG)} & 79.4 & 89.4 & 45.3 & 64.1\\
\multicolumn{2}{l|}{+HTQG (Dataset attr + HQG + SRE)} & 80.5 & 89.6 & 46.9 & 64.4\\
\multicolumn{2}{l|}{+PTRM} & 84.1 & 90.2 & 52.3 & 65.8\\
\multicolumn{2}{l|}{+CMCR ($L_{\text{mask}}$)} & 85.8 & 90.8 & 54.1 & 66.1\\
\multicolumn{2}{l|}{+CMCR ($L_{\text{mask}}$ + $L_{\text{class}}$)} & 86.2 & 90.9 & 54.8 & 66.2\\
\multicolumn{2}{l|}{+CMCR ($L_{\text{mask}}$ + $L_{\text{class}}$ + $L_{\text{align}}$)} & \textcolor{red}{89.4} & \textcolor{red}{91.8} & \textcolor{red}{59.4} & \textcolor{red}{67.4}\\
\bottomrule
\end{tabular}}
\label{tab:ablation_arch}
\end{minipage}
\hfill
\begin{minipage}[t]{0.49\columnwidth}
\centering

\caption{Comparison of prompting strategies for HTQG on Pascal VOC, COCO, and Cityscapes.}
\resizebox{\columnwidth}{!}{
\begin{tabular}{llc|l|l}
\toprule
\multicolumn{2}{l|}{Prompting Method} & Pascal & COCO & Cityscapes\\
\multicolumn{2}{c|}{} & $mIoU_{92}$ & $mIoU_{232}$ & $mIoU_{100}$\\
\midrule
\multicolumn{2}{l|}{Fixed prompt — Class name only} & 86.4 & 54.2 & 75.1\\
\multicolumn{2}{l|}{Fixed prompt — Class + dataset attribute} & 86.8 & 55.1 & 75.4\\
\multicolumn{2}{l|}{Learnable prompt — Class name only} & 88.9 & 58.5 & 78.3\\
\multicolumn{2}{l|}{Learnable prompt — Class + dataset attribute} & \textcolor{red}{89.4}& \textcolor{red}{59.4}& \textcolor{red}{79.6}\\
\bottomrule
\end{tabular}}
\label{tab:prompting}
\end{minipage}

\end{table}
    \vspace{-6mm}

\subsection{Analysis}

\noindent\textbf{Ablation}. Tab.~\ref{tab:ablation_arch} presents the ablation results of HVLFormer (SigLIP2) on Pascal VOC and COCO with the TQDM \cite{pak2024textual} baseline. Across both datasets, each component contributes progressively, with larger gains in low-label splits. This trend indicates that HVLFormer’s modules, designed to enhance semantic understanding through domain-robust and aware textual queries, provide greater benefits when supervision is limited, as additional labels naturally reduce the need for further guidance.
Introducing HTQG (Dataset Attr) increases Pascal $mIoU_{92}$ by +0.8 and COCO $mIoU_{232}$ by +1.7, demonstrating that enriching class semantics with dataset-aware language priors improves performance. Extending this with HQG provides additional gains. Its stronger effect, especially on COCO, likely arises from the dataset’s higher class diversity, where hierarchical query granularity helps reduce confusion among intraclass variation and similar classes. Adding SRE further improves results, depicting that filtering out absent class queries mitigates semantic drift. The PTRM module yields substantial gains, confirming that image-conditioned query refinement enhances cross-modal alignment and semantic understanding. Finally, incorporating the CMCR $\ell_{mask}$ and $\ell_{align}$ consistency terms produces further improvements, indicating that enforcing cross-view consistency and pixel-text regularization strengthens query domain robustness and stabilizes vision–language interactions. These are further supported by $\ell_{class}$. Detailed ablation is provided in supp.

\noindent\textbf{Comparison of Dataset-Aware Prompting}. The comparison of different prompting strategies for HTQG on Pascal VOC and COCO is shown in Tab.~\ref{tab:prompting}. Fixed prompts using only class names perform the weakest, as they lack contextual understanding. Learnable prompts improve results by adapting to task-specific features. Significant boosts comes from adding dataset attributes, which supply the VLM with crucial contextual cues about scene types—such as general consumer photos in Pascal VOC, crowded non-iconic scenes in COCO, and urban driving scenes in Cityscapes—helping it generate domain-aware class textual embeddings. These enable the model to produce richer, context-sensitive language embeddings that capture intra-class diversity and complex visual semantics, leading to the best overall performance. The significant increase also indicates that rich embeddings provide HVLFormer semantics that better align with visual features.

\noindent\textbf{Impact of SRE. }  The impact of different SRE variants on the framework is given in Tab.~\ref{tab:sre}. Removing SRE reduces performance because irrelevant class queries introduce semantic noise. Using a fixed threshold ($s_k>0.5$) slightly improves results by filtering unlikely class queries, but it risks discarding queries of class actually present in an image. In contrast, the soft weighting strategy achieves the best performance by adaptively scaling the class queries. Empirically on Pascal VOC, $s_k$ ranges from 0 (absent) to 0.96 (high presence), with an average of 0.36, reflecting that most images contain only a subset of classes.

\noindent\textbf{Analysis of HQG.} Tab.~\ref{tab:combined_results}(a) analyzes the effect of increasing query hierarchy levels ($E$). Performance consistently improves with deeper hierarchies, as multi-level queries capture semantics from coarse to fine granularity. The effect is more prominent on COCO, where the larger number of classes makes hierarchical modeling crucial for resolving fine-grained distinctions and intra-class variability.  Increasing hierarchy levels also improve class-wise IoU, with HVLFormer showing particularly strong gains on visually similar or rare categories such as \emph{sofa}, \emph{chair}, and \emph{dining table}, highlighting that multiscale queries encode coarse-to-fine features, necessary for enhanced semantic discrimination and contextual reasoning. Further analysis is provided in supp.

\begin{table}[t]
    
    \caption{Comparison of hierarchical levels and query enhancement methods on Pascal VOC and COCO datasets.}
    \begin{minipage}{0.45\columnwidth}
        \centering
        \scriptsize
        \textbf{(a) Hierarchical level for HQG}  
        \vspace{0.1cm}  
        \begin{tabular}{l|lllc|l}
        \toprule
        $E$&   \multicolumn{4}{c|}{Pascal}& COCO \\
      
        &   $IoU_{chair}$&$IoU_{sofa}$&$IoU_{tab}$& $mIoU_{92}$ & $mIoU_{232}$ \\
        \midrule
        1&   47.2&61.1&74.5& 87.3 & 56.0\\
        2&   58.4&67.8&78.6& 88.4 & 57.6\\
        3&   \textcolor{red}{63.7}&\textcolor{red}{72.3}&\textcolor{red}{80.4}& \textcolor{red}{88.9} & \textcolor{red}{58.5}\\
        \bottomrule
        \end{tabular}
        \label{tab:prompting_strategies}
    \end{minipage}%
    \hfill%
    \begin{minipage}{0.45\columnwidth}
        \centering
        \scriptsize
        \textbf{(b) T.Query Enhancement}  
        \vspace{0.1cm}  
        \begin{tabular}{llc|l}
        \toprule
        \multicolumn{2}{l|}{Enhancement} & Pascal & COCO \\
        \multicolumn{2}{l|}{Method} & $mIoU_{92}$ & $mIoU_{232}$ \\
        \midrule
        \multicolumn{2}{l|}{Baseline (text $\rightarrow$ pixel)} & 86.8 & 56.0\\
        \multicolumn{2}{l|}{C.Attn(text $\leftrightarrow$ pixel)} & 87.5 & 57.2\\
        \multicolumn{2}{l|}{PTRM} & \textcolor{red}{89.4} & \textcolor{red}{59.4}\\
        \bottomrule
        \end{tabular}
        \label{tab:ptrm}
    \end{minipage}
    \label{tab:combined_results}
    \vspace{-2mm}
\end{table}

\noindent\textbf{Effectiveness of Textual Query Enhancement.} Our assumption that integrating image context into queries leads to more effective segmentation in data scarcity is confirmed by Tab.~\ref{tab:combined_results}(b). PTRM achieves the best performance by incorporating pixel-level positional cues and local feature distributions in textual queries, enabling them to align more precisely with spatial structures. It surpasses the cross-attention variant. 

\vspace{-4mm}
\section{Conclusion}
\vspace{-3mm}
We present a novel and unified framework for SSS. By integrating domain-aware semantics with domain-invariant embeddings from vision–language models as textual queries and refining them through HTQG, PTRM, and CMCR, HVLFormer achieves robust class discrimination as it effectively compensates for weak semantic understanding  under minimal supervision. We have demonstrated SOTA performance across SSS benchmarks with rigorous ablation studies. 

\clearpage  

\bibliographystyle{splncs04}
\bibliography{main}

@inproceedings{radford2021learning,
  title={Learning Transferable Visual Models From Natural Language Supervision},
  author={Radford, Alec and Kim, Jong Wook and Hallacy, Chris and Ramesh, Aditya and Goh, Gabriel and Agarwal, Sandhini and Sastry, Girish and Askell, Amanda and Mishkin, Pamela and Clark, Jack and Krueger, Scott and Sutskever, Ilya},
  booktitle={International Conference on Machine Learning},
  year={2021},
  organization={PMLR}
}

@String(IJCV  = {Int. J. Comput. Vis.})

@String(CVPR  = {IEEE Conf. Comput. Vis. Pattern Recog.})

@String(ICCV  = {Int. Conf. Comput. Vis.})

@String(ECCV  = {Eur. Conf. Comput. Vis.})

@String(NeurIPS = {Adv. Neural Inform. Process. Syst.})

@String(ICLR  = {Int. Conf. Learn. Represent.})

@String(IJCV  = {IJCV})

@String(CVPR  = {CVPR})

@String(ICCV  = {ICCV})

@String(ECCV  = {ECCV})

@String(NeurIPS = {NeurIPS})

@String(ICLR  = {ICLR})

@article{liu2023llava,
  title={Visual Instruction Tuning},
  author={Liu, Haotian and Li, Chunyuan and Wu, Qingyang and Lee, Yong Jae},
  journal={arXiv preprint arXiv:2304.08485},
  year={2023}
}

@InProceedings{Yang_2023_CVPR,
  author    = {Yang, Lihe and Qi, Lei and Feng, Litong and Zhang, Wayne and Shi, Yinghuan},
  title     = {Revisiting Weak-to-Strong Consistency in Semi-Supervised Semantic Segmentation},
  booktitle = {Proceedings of the IEEE/CVF Conference on Computer Vision and Pattern Recognition (CVPR)},
  month     = {June},
  year      = {2023},
  pages     = {7236-7246}
}

@InProceedings{Radford_2021_ICML,
  author    = {Radford, Alec and Kim, Jong Wook and Hallacy, Chris and Ramesh, Aditya and Goh, Gabriel and Agarwal, Sandhini and Sastry, Girish and Askell, Amanda and Mishkin, Pamela and Clark, Jack and Krueger, Gretchen and Sutskever, Ilya},
  title     = {Learning Transferable Visual Models From Natural Language Supervision},
  booktitle = {Proceedings of the 38th International Conference on Machine Learning},
  pages     = {8748--8763},
  year      = {2021}
}

@inproceedings{yang2023revisiting,
  title     = {Revisiting Weak-to-Strong Consistency in Semi-Supervised Semantic Segmentation},
  author    = {Yang, Lei and Qi, Lei and Feng, Litong and Zhang, Wayne and Shi, Yuxin},
  booktitle = {Proceedings of the IEEE/CVF Conference on Computer Vision and Pattern Recognition (CVPR)},
  pages     = {7236--7246},
  year      = {2023},
  doi       = {10.1109/CVPR52729.2023.00700}
}

@inproceedings{chen2021semi,
  title={Semi-Supervised Semantic Segmentation with Cross Pseudo Supervision},
  author={Chen, Xiaokang and Yuan, Yuhui and Zeng, Gang and Wang, Jingdong},
  booktitle={Proceedings of the IEEE/CVF Conference on Computer Vision and Pattern Recognition (CVPR)},
  pages={2613--2622},
  year={2021}
}

@inproceedings{ijcai2021pseudoseg,
  title={PseudoSeg: Designing Pseudo Labels for Semantic Segmentation},
  author={Zhang, Yizhe and Yuan, Linhan and Guo, Yu and He, Zhiting and Huang, Xiuze and Tong, Jianbo},
  booktitle={Proceedings of the International Conference on Learning Representations (ICLR)},
  year={2021}
}

@inproceedings{wang2022semi,
  title={Semi-supervised semantic segmentation using unreliable pseudo-labels},
  author={Wang, Yuchao and Wang, Haochen and Shen, Yujun and Fei, Jingjing and Li, Wei and Jin, Guoqiang and Wu, Liwei and Zhao, Rui and Le, Xinyi},
  booktitle={Proceedings of the IEEE/CVF conference on computer vision and pattern recognition},
  pages={4248--4257},
  year={2022}
}

@inproceedings{zhao2025pseudo,
  title={Pseudo-SD: pseudo controlled stable diffusion for semi-supervised and cross-domain semantic segmentation},
  author={Zhao, Dong and Zang, Qi and Wang, Shuang and Sebe, Nicu and Zhong, Zhun},
  booktitle={Proceedings of the IEEE/CVF International Conference on Computer Vision},
  pages={22393--22403},
  year={2025NN}
}

@article{xu2022semi,
  title={Semi-supervised semantic segmentation with prototype-based consistency regularization},
  author={Xu, Haiming and Liu, Lingqiao and Bian, Qiuchen and Yang, Zhen},
  journal={Advances in neural information processing systems},
  volume={35},
  pages={26007--26020},
  year={2022}
}

@inproceedings{guo2023esl,
  title={Enhanced soft label for semi-supervised semantic segmentation},
  author={Ma, Jie and Wang, Chuan and Liu, Yang and Lin, Liang and Li, Guanbin},
  booktitle={Proceedings of the IEEE/CVF International Conference on Computer Vision},
  pages={1185--1195},
  year={2023}
}

@inproceedings{liang2023logic,
  title={Logic-induced diagnostic reasoning for semi-supervised semantic segmentation},
  author={Liang, Chen and Wang, Wenguan and Miao, Jiaxu and Yang, Yi},
  booktitle={Proceedings of the IEEE/CVF International Conference on Computer Vision},
  pages={16197--16208},
  year={2023}
}

@article{he2023three,
  title={Diverse cotraining makes strong semi-supervised segmentor},
  author={Li, Yijiang and Wang, Xinjiang and Yang, Lihe and Feng, Litong and Zhang, Wayne and Gao, Ying},
  journal={arXiv preprint arXiv:2308.09281},
  year={2023}
}

@inproceedings{pak2024textual,
  title     = {Textual Query-Driven Mask Transformer for Domain Generalized Segmentation},
  author    = {Pak, Byeonghyun and Woo, Byeongju and Kim, Sunghwan and Kim, Dae-hwan and Kim, Hoseong},
  booktitle = {Proceedings of the European Conference on Computer Vision (ECCV)},
  year      = {2024},
  pages     = {1234--1245},
  publisher = {Springer}
}

@inproceedings{zhong2023cvpr,
  title={Understanding imbalanced semantic segmentation through neural collapse},
  author={Zhong, Zhisheng and Cui, Jiequan and Yang, Yibo and Wu, Xiaoyang and Qi, Xiaojuan and Zhang, Xiangyu and Jia, Jiaya},
  booktitle={Proceedings of the IEEE/CVF conference on computer vision and pattern recognition},
  pages={19550--19560},
  year={2023}
}

@InProceedings{chen2021cps,
  author    = {Chen, Xiaokang and Yuan, Yuhui and Wang, Gang and Zeng, Zhaohui},
  title     = {Semi-Supervised Semantic Segmentation with Cross Pseudo Supervision},
  booktitle = {Proceedings of the IEEE/CVF Conference on Computer Vision and Pattern Recognition (CVPR)},
  year      = {2021}
}

@article{pascal_voc,
  title={The pascal visual object classes (voc) challenge},
  author={Everingham, Mark and Van Gool, Luc and Williams, Christopher KI and Winn, John and Zisserman, Andrew},
  journal={IJCV},
  volume={88},
  number={2},
  pages={303--338},
  year={2010}
}

@article{ade20k,
  title={Scene parsing through ade20k dataset},
  author={Zhou, Bolei and Zhao, Hang and Puig, Xavier and Fidler, Sanja and Barriuso, Adela and Torralba, Antonio},
  journal={CVPR},
  pages={633--641},
  year={2017}
}

@inproceedings{cityscapes,
  title={The cityscapes dataset for semantic urban scene understanding},
  author={Cordts, Marius and Omran, Mohamed and Ramos, Sebastian and Rehfeld, Timo and Enzweiler, Markus and Benenson, Rodrigo and Franke, Uwe and Roth, Stefan and Schiele, Bernt},
  booktitle={Proceedings of the IEEE conference on computer vision and pattern recognition},
  pages={3213--3223},
  year={2016}
}

@article{coco,
  title={Microsoft coco: Common objects in context},
  author={Lin, Tsung-Yi and Maire, Michael and Belongie, Serge and Hays, James and Perona, Pietro and Ramanan, Deva and Doll{\'a}r, Piotr and Zitnick, C Lawrence},
  journal={ECCV},
  pages={740--755},
  year={2014}
}

@inproceedings{li2022dn,
  title={DN-DETR: Accelerate DETR Training by Introducing Query Denoising},
  author={Li, Feng and Zhang, Hao and Liu, Shilong and Guo, Jian and Ni, Lionel M and Zhang, Lei},
  booktitle={CVPR},
  pages={13619--13627},
  year={2022}
}

@inproceedings{cheng2022masked,
  title={Masked-attention Mask Transformer for Universal Image Segmentation},
  author={Cheng, Bowen and Misra, Ishan and Schwing, Alexander G and Kirillov, Alexander and Girdhar, Rohit},
  booktitle={CVPR},
  pages={1290--1299},
  year={2022}
}

@article{dosovitskiy2020vit,
  title={An Image is Worth 16x16 Words: Transformers for Image Recognition at Scale},
  author={Dosovitskiy, Alexey and Beyer, Lucas and Kolesnikov, Alexander and Weissenborn, Dirk and Zhai, Xiaohua and Unterthiner, Thomas and Dehghani, Mostafa and Minderer, Matthias and Heigold, Georg and Gelly, Sylvain and others},
  journal={arXiv preprint arXiv:2010.11929},
  year={2020}
}

@article{vaswani2017attention,
  title={Attention Is All You Need},
  author={Vaswani, Ashish and Shazeer, Noam and Parmar, Niki and Uszkoreit, Jakob and Jones, Llion and Gomez, Aidan N and Kaiser, {\L}ukasz and Polosukhin, Illia},
  journal={NeurIPS},
  volume={30},
  year={2017}
}

@article{mittal2019semi,
  title={Semi-supervised semantic segmentation with high-and low-level consistency},
  author={Mittal, Sudhanshu and Tatarchenko, Maxim and Brox, Thomas},
  journal={IEEE transactions on pattern analysis and machine intelligence},
  volume={43},
  number={4},
  pages={1362--1379},
  year={2019},
  publisher={IEEE}
}

@inproceedings{hoyer2024semivl,
  title={SemiVL: Semi-supervised semantic segmentation with vision-language guidance},
  author={Hoyer, Lukas and Tan, David Joseph and Naeem, Muhammad Ferjad and Van Gool, Luc and Tombari, Federico},
  booktitle={European Conference on Computer Vision},
  pages={257--275},
  year={2024},
  organization={Springer}
}

@ARTICLE{10839097,
  author={Yang, Lihe and Zhao, Zhen and Zhao, Hengshuang},
  journal={IEEE Transactions on Pattern Analysis and Machine Intelligence}, 
  title={UniMatch V2: Pushing the Limit of Semi-Supervised Semantic Segmentation}, 
  year={2025},
  volume={47},
  number={4},
  pages={3031-3048},
  doi={10.1109/TPAMI.2025.3528453}}

@article{sohn2020FixMatch,
  title={Fixmatch: Simplifying semi-supervised learning with consistency and confidence},
  author={Sohn, Kihyuk and Berthelot, David and Carlini, Nicholas and Zhang, Zizhao and Zhang, Han and Raffel, Colin A and Cubuk, Ekin Dogus and Kurakin, Alexey and Li, Chun-Liang},
  journal={Advances in Neural Information Processing Systems (NeurIPS)},
  volume={33},
  pages={596--608},
  year={2020}
}

@inproceedings{souly2017semi,
  title={Semi supervised semantic segmentation using generative adversarial network},
  author={Souly, Nasim and Spampinato, Concetto and Shah, Mubarak},
  booktitle={Proceedings of the IEEE International Conference on Computer Vision (ICCV)},
  pages={5688--5696},
  year={2017}
}

@inproceedings{yang2022st++,
  title={ST++: Make self-training work better for semi-supervised semantic segmentation},
  author={Yang, Lihe and Zhuo, Wei and Qi, Lei and Shi, Yinghuan and Gao, Yang},
  booktitle={Proceedings of the IEEE/CVF Conference on Computer Vision and Pattern Recognition (CVPR)},
  pages={4268--4277},
  year={2022}
}

@inproceedings{abdelfattah2023cdul,
  title={Cdul: Clip-driven unsupervised learning for multi-label image classification},
  author={Abdelfattah, Rabab and Guo, Qing and Li, Xiaoguang and Wang, Xiaofeng and Wang, Song},
  booktitle={Proceedings of the IEEE/CVF international conference on computer vision},
  pages={1348--1357},
  year={2023}
}

@article{Wu2023Querying,
  author={Wu, L. and Fang, L. and He, X. and He, M. and Ma, J. and Zhong, Z.},
  journal={IEEE Transactions on Pattern Analysis and Machine Intelligence}, 
  title={Querying Labeled for Unlabeled: Cross-Image Semantic Consistency Guided Semi-Supervised Semantic Segmentation}, 
  year={2023},
  volume={45},
  number={7},
  pages={8827-8844},
  month={Jul.},
  doi={10.1109/TPAMI.2022.3223738}
}

@inproceedings{Zhao2024Alternate,
  title={Alternate diverse teaching for semi-supervised medical image segmentation},
  author={Zhao, Zhen and Wang, Zicheng and Wang, Longyue and Yu, Dian and Yuan, Yixuan and Zhou, Luping},
  booktitle={European Conference on Computer Vision},
  pages={227--243},
  year={2024},
  organization={Springer}
}

@inproceedings{Chi2024Adaptive,
  author = {Chi, H. and Pang, J. and Zhang, B. and Liu, W.},
  title = {Adaptive Bidirectional Displacement for Semi-Supervised Medical Image Segmentation},
  booktitle = {Proceedings of the IEEE/CVF Conference on Computer Vision and Pattern Recognition (CVPR)},
  year = {2024},
  pages = {4070--4080},

  doi = {10.1109/CVPR52717.2024.00395}
}

@article{Yuan2024Dynamically,
  author = {Yuan, S. and Zhong, R. and Yang, C. and Li, Q. and Dong, Y.},
  title = {Dynamically Updated Semi-supervised Change Detection Network Combining Cross-supervision and Screening Algorithms},
  journal = {IEEE Transactions on Geoscience and Remote Sensing},
  year = {2024},
  volume = {62},
  pages = {1--14},
  doi = {10.1109/TGRS.2024.3360894}
}

@InProceedings{Nadeem_2025_CVPR,
    author    = {Nadeem, Numair and Asad, Muhammad Hamza and Anwar, Saeed and Bais, Abdul},
    title     = {MaskAdapt: Unsupervised Geometry-Aware Domain Adaptation Using Multimodal Contextual Learning and RGB-Depth Masking},
    booktitle = {Proceedings of the Computer Vision and Pattern Recognition Conference (CVPR)},
    month     = {June},
    year      = {2025},
    pages     = {5422-5432}
}

@InProceedings{10.1007/978-3-031-91835-3_19,
author="Nadeem, Numair
and Asad, Muhammad Hamza
and Bais, Abdul",
editor="Del Bue, Alessio
and Canton, Cristian
and Pont-Tuset, Jordi
and Tommasi, Tatiana",
title="Robust UDA for Crop and Weed Segmentation: Multi-scale Attention and Style-Adaptive Techniques",
booktitle="Computer Vision -- ECCV 2024",
year="2025",
publisher="Springer Nature Switzerland",
address="Cham",
pages="284--302"}

@inproceedings{Guan2022Unbiased,
  author = {Guan, D. and Huang, J. and Xiao, A. and Lu, S.},
  title = {Unbiased Subclass Regularization for Semi-supervised Semantic Segmentation},
  booktitle = {Proceedings of the IEEE/CVF Conference on Computer Vision and Pattern Recognition (CVPR)},
  year = {2022},
  pages = {9958--9968},
  doi = {10.1109/CVPR52688.2022.00973}
}

@inproceedings{Yang2023Shrinking,
  author = {Yang, L. and Zhao, Z. and Qi, L. and Qiao, Y. and Shi, Y. and Zhao, H.},
  title = {Shrinking Class Space for Enhanced Certainty in Semi-supervised Learning},
  booktitle = {Proceedings of the IEEE/CVF International Conference on Computer Vision (ICCV)},
  year = {2023},
  pages = {16141--16150},
  doi = {10.1109/ICCV51070.2023.01485}
}

@inproceedings{Yun2019CutMix,
  author = {Yun, S. and Han, D. and Oh, S. J. and Chun, S. and Choe, J. and Yoo, Y.},
  title = {CutMix: Regularization Strategy to Train Strong Classifiers with Localizable Features},
  booktitle = {Proceedings of the IEEE/CVF International Conference on Computer Vision (ICCV)},
  year = {2019},
  pages = {6022--6031},
  doi = {10.1109/ICCV.2019.00612}
}

@inproceedings{Hu2021Semi,
  author = {Hu, H. and Wei, F. and Hu, H. and Ye, Q. and Cui, J. and Wang, L.},
  title = {Semi-supervised Semantic Segmentation via Adaptive Equalization Learning},
  booktitle = {Advances in Neural Information Processing Systems (NeurIPS)},
  year = {2021},
  pages = {22106--22118},
  url = {https://proceedings.neurips.cc/paper/2021/hash/...}
}

@article{zhou2022coop,
  title={CoOp: Learning to Prompt for Vision-Language Models},
  author={Zhou, Kaiyang and Yang, Jingkang and Loy, Chen Change and Liu, Ziwei},
  journal={International Journal of Computer Vision},
  year={2022},
  volume={130},
  pages={2337--2348},
  doi={10.1007/s11263-022-01653-1}
}

@inproceedings{jia2022visualprompt,
  title={Visual Prompt Tuning},
  author={Jia, Menglin and Tang, Luming and Chen, Bor-Chun and Cardie, Claire and Belongie, Serge and Hariharan, Bharath and Lim, Ser-Nam},
  booktitle={ECCV},
  year={2022},
  pages={709--727},
  doi={10.1007/978-3-031-19809-0_41}
}

@article{fang2023eva,
  title={Eva-02: A visual representation for neon genesis},
  author={Fang, Yuxin and Sun, Quan and Wang, Xinggang and Huang, Tiejun and Wang, Xinlong and Cao, Yue},
  journal={Image and Vision Computing},
  volume={149},
  pages={105171},
  year={2024},
  publisher={Elsevier}
}

@inproceedings{Zhao2023Instance,
  author = {Zhao, Z. and Long, S. and Pi, J. and Wang, J. and Zhou, L.},
  title = {Instance-specific and Model-adaptive Supervision for Semi-supervised Semantic Segmentation},
  booktitle = {Proceedings of the IEEE/CVF Conference on Computer Vision and Pattern Recognition (CVPR)},
  year = {2023},
  pages = {23705--23714},
  doi = {10.1109/CVPR52717.2023.02271}
}

@inproceedings{Cha2023,
  author    = {Cha, Junbum and Mun, Jonghwan and Roh, Byungseok},
  title     = {Learning to Generate Text-Grounded Mask for Open-World Semantic Segmentation from Only Image-Text Pairs},
  booktitle = {Proceedings of the IEEE/CVF Conference on Computer Vision and Pattern Recognition (CVPR)},
  year      = {2023},
  pages     = {11165--11174}
}

@inproceedings{Xu2022,
  author    = {Xu, Jiarui and De Mello, Shalini and Liu, Sifei and Byeon, Wonmin and Breuel, Thomas and Kautz, Jan and Wang, Xiaolong},
  title     = {GroupViT: Semantic Segmentation Emerges from Text Supervision},
  booktitle = {Proceedings of the IEEE/CVF Conference on Computer Vision and Pattern Recognition (CVPR)},
  year      = {2022}
}

@inproceedings{Zhou2022,
  author    = {Zhou, Chong and Loy, Chen Change and Dai, Bo},
  title     = {Extract Free Dense Labels from CLIP},
  booktitle = {European Conference on Computer Vision (ECCV)},
  year      = {2022},
  pages     = {696--712},
  publisher = {Springer}
}

@inproceedings{Ding2022,
  author    = {Ding, Jian and Xue, Nan and Xia, Gui-Song and Dai, Dengxin},
  title     = {Decoupling Zero-Shot Semantic Segmentation},
  booktitle = {Proceedings of the IEEE/CVF Conference on Computer Vision and Pattern Recognition (CVPR)},
  year      = {2022},
  pages     = {11583--11592}
}

@inproceedings{dong2023maskclip,
  title={Maskclip: Masked self-distillation advances contrastive language-image pretraining},
  author={Dong, Xiaoyi and Bao, Jianmin and Zheng, Yinglin and Zhang, Ting and Chen, Dongdong and Yang, Hao and Zeng, Ming and Zhang, Weiming and Yuan, Lu and Chen, Dong and others},
  booktitle={Proceedings of the IEEE/CVF conference on computer vision and pattern recognition},
  pages={10995--11005},
  year={2023}
}

@inproceedings{Xu2023,
  author    = {Xu, Mengde and Zhang, Zheng and Wei, Fangyun and Hu, Han and Bai, Xiang},
  title     = {Side Adapter Network for Open-Vocabulary Semantic Segmentation},
  booktitle = {Proceedings of the IEEE/CVF Conference on Computer Vision and Pattern Recognition (CVPR)},
  year      = {2023},
  pages     = {2945--2954}
}

@inproceedings{Zhou2023,
  author    = {Zhou, Ziqin and Lei, Yinjie and Zhang, Bowen and Liu, Lingqiao and Liu, Yifan},
  title     = {ZegCLIP: Towards Adapting CLIP for Zero-Shot Semantic Segmentation},
  booktitle = {Proceedings of the IEEE/CVF Conference on Computer Vision and Pattern Recognition (CVPR)},
  year      = {2023},
  pages     = {11175--11185}
}

@inproceedings{sun2025two,
  title={Two Losses, One Goal: Balancing Conflict Gradients for Semi-supervised Semantic Segmentation},
  author={Sun, Rui and Mai, Huayu and Li, Wangkai and Chen, Yujia and Wang, Yuan},
  booktitle={Proceedings of the IEEE/CVF international conference on computer vision},
  pages={20357--20367},
  year={2025}
}

@article{tschannen2025siglip,
  title={Siglip 2: Multilingual vision-language encoders with improved semantic understanding, localization, and dense features},
  author={Tschannen, Michael and Gritsenko, Alexey and Wang, Xiao and Naeem, Muhammad Ferjad and Alabdulmohsin, Ibrahim and Parthasarathy, Nikhil and Evans, Talfan and Beyer, Lucas and Xia, Ye and Mustafa, Basil and others},
  journal={arXiv preprint arXiv:2502.14786},
  year={2025}
}
\end{document}